\pdfoutput=1

\documentclass[11pt]{article}

\usepackage[preprint]{acl}

\usepackage{times}
\usepackage{latexsym}
\usepackage{url}
\usepackage[T1]{fontenc}

\usepackage[utf8]{inputenc}

\usepackage{microtype}

\usepackage{threeparttable}
\usepackage{adjustbox}

\usepackage{inconsolata}

\usepackage{graphicx}

\usepackage{framed}
\usepackage{listings}

\lstdefinestyle{promptstyle}{
  basicstyle=\ttfamily\small,
  breaklines=true,
  frame=single,
  backgroundcolor=\color{gray!5},
  captionpos=b
}

\usepackage{amsmath,amsfonts,bm}
\usepackage{booktabs} 
\usepackage{url}
\usepackage{graphicx}
\usepackage{array}
\usepackage{color}
\usepackage{makecell}
\usepackage{multirow}
\usepackage{xspace}
\usepackage{colortbl}
\usepackage{subcaption}
\usepackage{algpseudocode}
\usepackage[noend,ruled]{algorithm2e}
\usepackage{setspace}
\usepackage{varwidth}
\usepackage{enumitem}
\usepackage{mdframed}
\usepackage{centernot}
\usepackage{arydshln}

\usepackage{soul}
\usepackage{pifont}
\usepackage{graphics}

\usepackage{wrapfig}

\usepackage{cuted,tcolorbox,lipsum}

\newcommand{\circled}[1]{\textcircled{\raisebox{-0.9pt}{#1}}}

\newcommand{\bigcircled}[1]{\textcircled{\raisebox{-0.9pt}{\small{#1}}}}

\newcommand{\start}[1]{\vspace{.3mm}\noindent{{\bf #1}.}}

\definecolor{amber}{rgb}{1.0, 0.75, 0.0}
\definecolor{applegreen}{rgb}{0.55, 0.71, 0.0}
\definecolor{treegreen}{rgb}{0.13, 0.54, 0.23}
\definecolor{LightCyan}{rgb}{0.88,1,1}
\definecolor{LightBlue}{RGB}{171, 227, 235}

\definecolor{green2}{HTML}{BFD8B6}
\definecolor{green3}{HTML}{E7F0E5}
\definecolor{greenarrow}{HTML}{1DB100}
\definecolor{red3}{HTML}{C82506}

\definecolor{gred}{RGB}{255,102,102}
\definecolor{gblue}{RGB}{51,102,255}
\definecolor{gyellow}{RGB}{244,180,0}
\definecolor{ggreen}{RGB}{15,157,88}
\definecolor{ggrey}{RGB}{115,115,115}
\definecolor{na}{gray}{0.9}

\definecolor{textRed}{RGB}{157,0,23}
\definecolor{textYellow}{RGB}{166,119,54}
\definecolor{textGreen}{RGB}{58,110,38}
\definecolor{textBlue}{RGB}{39,71,156}

\definecolor{LightYellow}{RGB}{255,250,208}
\definecolor{LightGreen}{RGB}{194,255,192}
\definecolor{LightBlue}{RGB}{187,236,251}
\definecolor{LightPurple}{RGB}{224,223,255}
\definecolor{LightGrey}{RGB}{225,225,225}

\definecolor{Grey}{RGB}{150,150,150}

\definecolor{OrangeRed}{rgb}{1.0, 0.27, 0.0}
\definecolor{midnightgreen}{rgb}{0.0, 0.29, 0.33}
\definecolor{darkgreen}{rgb}{0.0, 0.42, 0.24}

\definecolor{diagramRed}{RGB}{246,193,193}
\definecolor{diagramPurple}{RGB}{224,224,253}
\definecolor{diagramOrange}{RGB}{244,222,176}

\title{An Empirical Study on Strong-Weak Model Collaboration for \\ Repo-level Code Generation}

\author{Shubham Gandhi, Atharva Naik, Yiqing Xie, Carolyn Rose \\
  Language Technologies Institute, Carnegie Mellon University \\
  \texttt{\{srgandhi, arnaik, yiqingxi, cp3a\}@andrew.cmu.edu}}

\begin{document}
\maketitle
\begin{abstract}
We study cost-efficient collaboration between strong and weak language models for repository-level code generation, where the weak model handles simpler tasks at lower cost, and the most challenging tasks are delegated to the strong model.
While many works propose architectures for this task, few analyze performance relative to cost. 
We evaluate a broad spectrum of collaboration strategies: context-based, pipeline-based, and dynamic, on GitHub issue resolution.
Our most effective collaborative strategy achieves equivalent performance to the strong model while reducing the cost by 40\%.
Based on our findings, we offer actionable guidelines for choosing collaboration strategies under varying budget and performance constraints.
Our results show that strong–weak collaboration substantially boosts the weak model’s performance at a fraction of the cost, pipeline and context-based methods being most efficient. We release the code \footnote{\url{https://github.com/shubhamrgandhi/codegen-strong-weak-collab}} for our work.


\end{abstract}

\begin{figure*}[!htbp]
    \centering
    \vspace{-1cm}
    \includegraphics[width=1\textwidth]{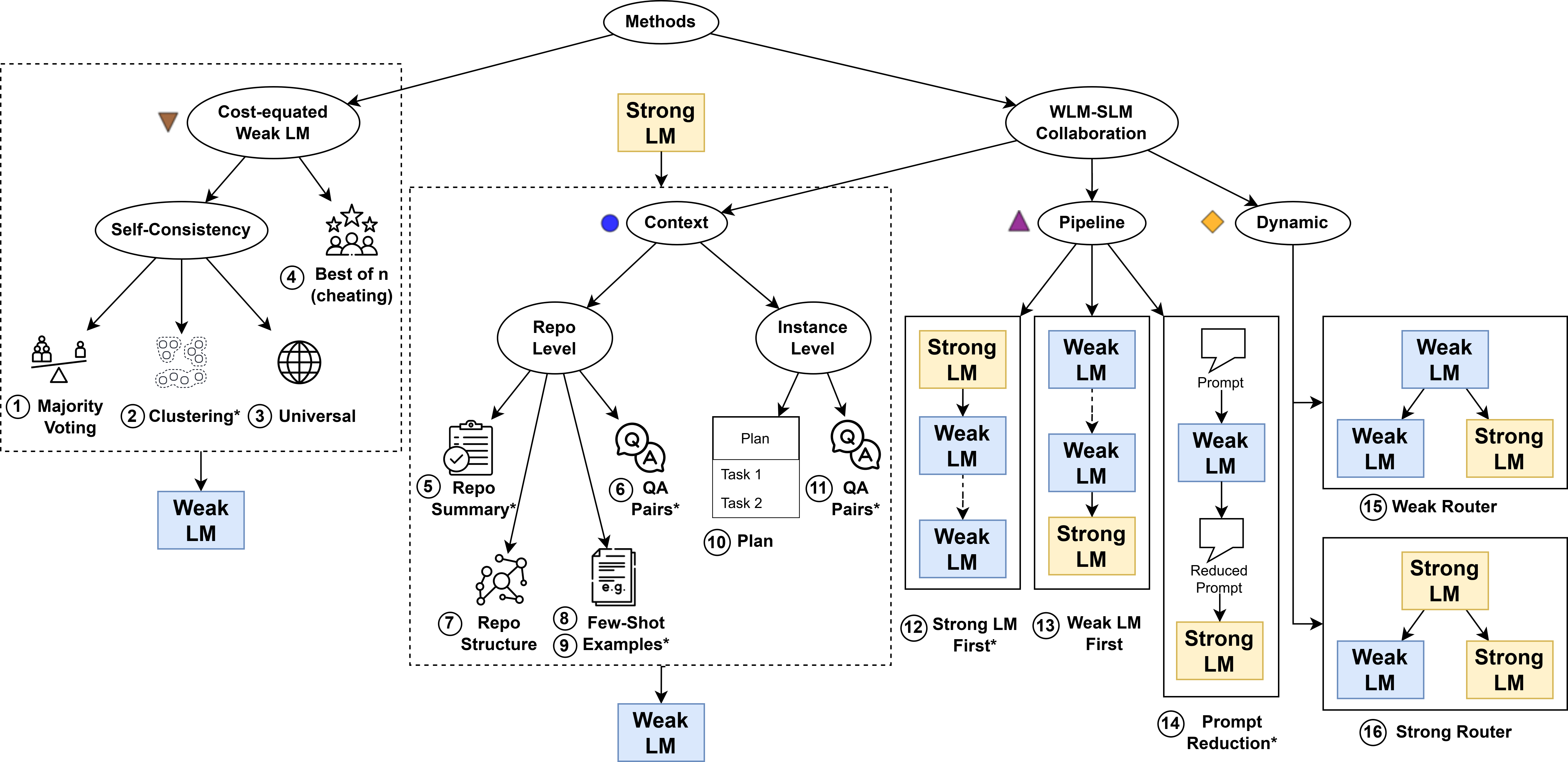}
    \caption{Taxonomy of the 14 techniques studied. * denotes methods newly proposed or adapted in this study. We categorize them into cost-equated weak-only, context-based, pipeline-based, and dynamic collaboration methods.}
    \label{fig:taxonomy}
\end{figure*}

\section{Introduction}
\label{sec:introduction}






Recent advances in large language models (LLMs) have demonstrated impressive capabilities across complex reasoning and generation tasks. However, benchmark gains are often prioritized over practical deployment concerns, leading to systems that rely on repeated calls to expensive proprietary LLMs. 
For example, SWE-Agent \cite{yang2024sweagentagentcomputerinterfacesenable} caps each run at \$4
~\cite{kapoor2024aiagentsmatter}
, making even modest-scale evaluations prohibitively expensive.
In contrast, systems such as Retrieval-Augmented Generation (RAG) can reduce cost by 99\%
~\cite{jin2025llmsllmbasedagentssoftware}, motivating a critical question: \textit{How can we design effective yet cost-efficient LLM systems?} 
Retrieval-Augmented Code Generation (RACG)~\cite{wang-etal-2025-coderag} systems like Agentless~\cite{agentless} and Agentless Lite~\cite{AgentlessLite2025} try to address this by selectively retrieving relevant context for generation, significantly reducing costs (\$0.25 / run).
However, varying task complexity  
necessitates hybrid setups where weak LLMs economically handle simpler tasks and escalate the rest. Yet, designing such systems is non-trivial because weak LLMs struggle with complexity, and overusing strong LLMs negates cost benefits. 

While prior work explored strong-weak collaboration in NLP via 
LLM cascades \cite{chen2023frugalgptuselargelanguage, zhang2023ecoassistantusingllmassistant, yue2024largelanguagemodelcascades}, context augmentation \cite{chen2025codesteersymbolicaugmentedlanguagemodels}, ensembling \cite{zhang2024smalllanguagemodelsneed, chen2024improvinglargemodelssmall}, etc, a systematic analysis for repository level code generation remains unexplored.

To bridge this gap, 
we present a three-pronged taxonomy with 12 diverse strong-weak collaboration methods, including static context augmentation (e.g., Few-Shot examples, FAQs, Planning), pipeline division (e.g., Strong LM First, Prompt Reduction), and dynamic collaboration (e.g., Routing). 
Our taxonomy (Figure~\ref{fig:taxonomy}) draws on prior work along with tailored adaptations for code generation. We evaluate these methods across both API-based and hybrid (API + open-source) LLM pairs on the SWE-Bench Lite~\cite{jimenez2024swebench} benchmark.
Our goal is to empirically characterize the cost-performance trade-offs of these methods and identify best practices under different budget and accuracy constraints. While issue-solving serves as a grounded and challenging testbed, our broader focus is on the general design space of strong–weak collaboration.
 
Our findings show that pipeline and context-based strategies offer the best average efficiency, while weak-only baselines consistently underperform. Yet, we find that examining average efficiency does not always predict optimal performance for a given cost. Through cost-performance curves, we answer the question: \textit{What is the best method given a performance requirement and budget constraint?} 
Our best strategy improves the weak LLM's performance by $\sim$62\%, almost matching the strong LLM, at $\sim$60\% cost. 
Overall, we provide actionable principles for designing cost-efficient, collaborative LLM systems for complex code generation tasks in the future.

\section{Related Work}
\label{sec:related_work}


Strong LLMs
achieve high performance but are often too expensive to use at scale. To tackle this, several lines of work explore cost-efficient alternatives.  Self-consistency \cite{wang2023selfconsistencyimproveschainthought, chen2023universalselfconsistencylargelanguage}  aggregates multiple generations from the same LLM to improve performance. Other approaches rely on strong–weak collaboration.~\citet{amayuelas2025selfresourceallocationmultiagentllm, wang2024planningnaturallanguageimproves} leverage a strong LM's superior reasoning capabilities to generate a plan for execution by a cheaper weak LM.
Routing-based methods like FrugalGPT \cite{chen2023frugalgptuselargelanguage} dynamically select LLMs to optimize cost and quality. 
Similarly, LLM Cascades \cite{yue2024largelanguagemodelcascades} escalate to strong LMs only when weak LMs fail. 
Drawing from this, we curate a taxonomy of strong–weak collaboration strategies, including self-consistency, planning, LLM cascades (Weak LM First), routing among others, adapted to repository-level code generation. 
This allows us to empirically evaluate their cost-performance trade-offs in a unified framework.

\section{Methodology}
\label{sec:approach}


To conduct a systematic analysis, as shown in \autoref{fig:taxonomy}, we present a taxonomy of strong-weak collaboration methods for repo-level RACG, where the methods are either based on or inspired by prior studies \cite{chen2025reasoningerasurveylong, bai2024efficiencysystematicsurveyresourceefficient, xu2024surveyknowledgedistillationlarge, lu2024mergeensemblecooperatesurvey, chen2025harnessingmultiplelargelanguage}.


\start{Preliminary}
We conduct our experiments with Agentless-Lite~\cite{AgentlessLite2025}, a two-step repo-level RACG framework. We first apply a retriever to retrieve the top-$k$ relevant documents (e.g., files in a repository) based on their similarity score with the query $q$. Then we apply an LM to iteratively generate the code with the documents as context until it is in the correct code patch format.

\start{Cost-equated weak LM}
We first study several baselines that \emph{only} involve the weak LM~\cite{mcdonald-etal-2025-afford}. 
Self-consistency \citep{wang2023selfconsistencyimproveschainthought} enhances weak LMs by sampling $n$ diverse outputs, where $n$ $\approx$ Cost$_{strong}$ / Cost$_{weak}$ and selecting the most consistent one via:
\circled{1} \underline{Majority Voting ($SC_m$)},
\circled{2} \underline{Clustering ($SC_c$)} and 
\circled{3} \underline{Universal ($SC_u$)}.
We also experiment on a ``cheating'' method, \circled{4} \underline{Best of n}. Detailed explanations are included in Appendix \S \ref{sec:hyperparameters}

\start{Static Context Augmentation}
We experiment methods that leverage the strong LM's superior problem-understanding capabilities to augment the context and use the weak LM for iterative generation to reduce cost. 
We first study methods that prompt the strong LM to generate repo-level information 
to provide background information for code generation, including:
\circled{5} \underline{Repository Summary}, which generates a summary of the repository based on its \texttt{README} file and directory structure,
\circled{6} \underline{Repo-Level FAQs}, which is similar but generates a set of FAQs instead,
\circled{7} \underline{Repo Structure}, which summarizes the repository structure.
Here we use RepoGraph \citep{ouyang2025repographenhancingaisoftware} 
, which encodes the repository in a graph.
We also study few-shot examples for each repository, which can be constructed by: 
\circled{8} \underline{Few-shot (Random)} that randomly selects $k$ (input, successful strong LM output) pairs, and \circled{9} \underline{Few-shot (Similarity)}, which selects few-shot demos based on the problem statements' embedding similarity.
We also study methods that use the strong LM to generate instance-specific context. Compared to repo-level context, it typically requires a higher cost but provides more precise and example-relevant knowledge. We analyze:
\bigcircled{10} \underline{Planning}, which generates high-level planning for each instance ~\cite{amayuelas2025selfresourceallocationmultiagentllm, wang2024planningnaturallanguageimproves}, and
\bigcircled{11} \underline{Instance-Level QA pairs}, which generates a set of FAQs for each instance with the code generation query (e.g., issue description) and retrieved code as context.

\start{Pipeline Division} \label{sec:collab_pipeline}
These methods aim to reduce cost while maintaining performance by selectively calling the strong LM and weak LM sequentially as part of a hard-coded pipeline.
Specifically, we compare:
\bigcircled{12} \underline{Strong LM First}, which prompts the strong LM to make the first attempt and calls the weak LM to iteratively refine its solution until it is correctly formatted.
\bigcircled{13} \underline{Weak LM First}, where the weak LM first makes $n$ attempts to solve the issue, following model cascades~\cite{chen2023frugalgptuselargelanguage, zhang2023ecoassistantusingllmassistant, yue2024largelanguagemodelcascades}. If it fails to generate a valid patch, the strong LM makes one attempt.
Based on previous study's conclusion that weak LMs have comparable performance to strong LMs on localization~\cite{agentless}, 
we present \bigcircled{14} \underline{Prompt Reduction}, where we perform a preliminary call to the weak LM to reduce the code context by removing irrelevant code, thus reducing the overall context to be passed to the strong LM.

\start{Dynamic Collaboration}
Unlike \S \ref{sec:collab_pipeline}, where the pipeline is hard-coded, we allow the decision to use the strong or weak model to be made dynamically during inference. Specifically, we invoke a router~\cite{chen2023frugalgptuselargelanguage} to decide if the problem is \textit{simple} or \textit{complex} and appropriately delegate it to the weak or strong LM. We evaluate both weak LMs and strong LMs as routers and denote them as \bigcircled{15} \underline{Weak Router} and \bigcircled{16} \underline{Strong Router}.

\section{Experiments}

\start{Models}
We consider strong-weak model pairs that have a noticeable gap in performance and cost, and the more expensive model performs better: \textbf{Strong LMs}: O3-mini, O4-mini~\cite{openai2025o3mini} and GPT-4o-mini~\cite{openai2024gpt4o};
\textbf{Weak LMs}: GPT-4o-mini and Qwen2.5-Coder series~\cite{hui2024qwen25codertechnicalreport}.


\start{Dataset and Agentic Framework}
We conduct all experiments on the SWE-bench Lite subset \cite{jimenez2024swebench} which has 300 issues from 11 Python repositories. We run the Agentless Lite framework \cite{AgentlessLite2025}
once per instance for our experiments, using \texttt{voyage-code-3}\footnote{\url{https://docs.voyageai.com/docs/embeddings}} for retrieval, techniques mentioned in \S\ref{sec:approach} for generation (Details in Appendix \S\ref{sec:hyperparameters}).

\start{Evaluation Metrics}
The following metrics are reported for each method: (1) \textit{Resolution Rate}: Proportion of instances for which the generated patch successfully resolves the issue; and
(2) \textit{Cost}: Total generation cost (\$) 
including additional method costs, but excluding retrieval cost, which remains constant across all methods. 

\begin{figure}
    \centering
    \includegraphics[width=1\linewidth]{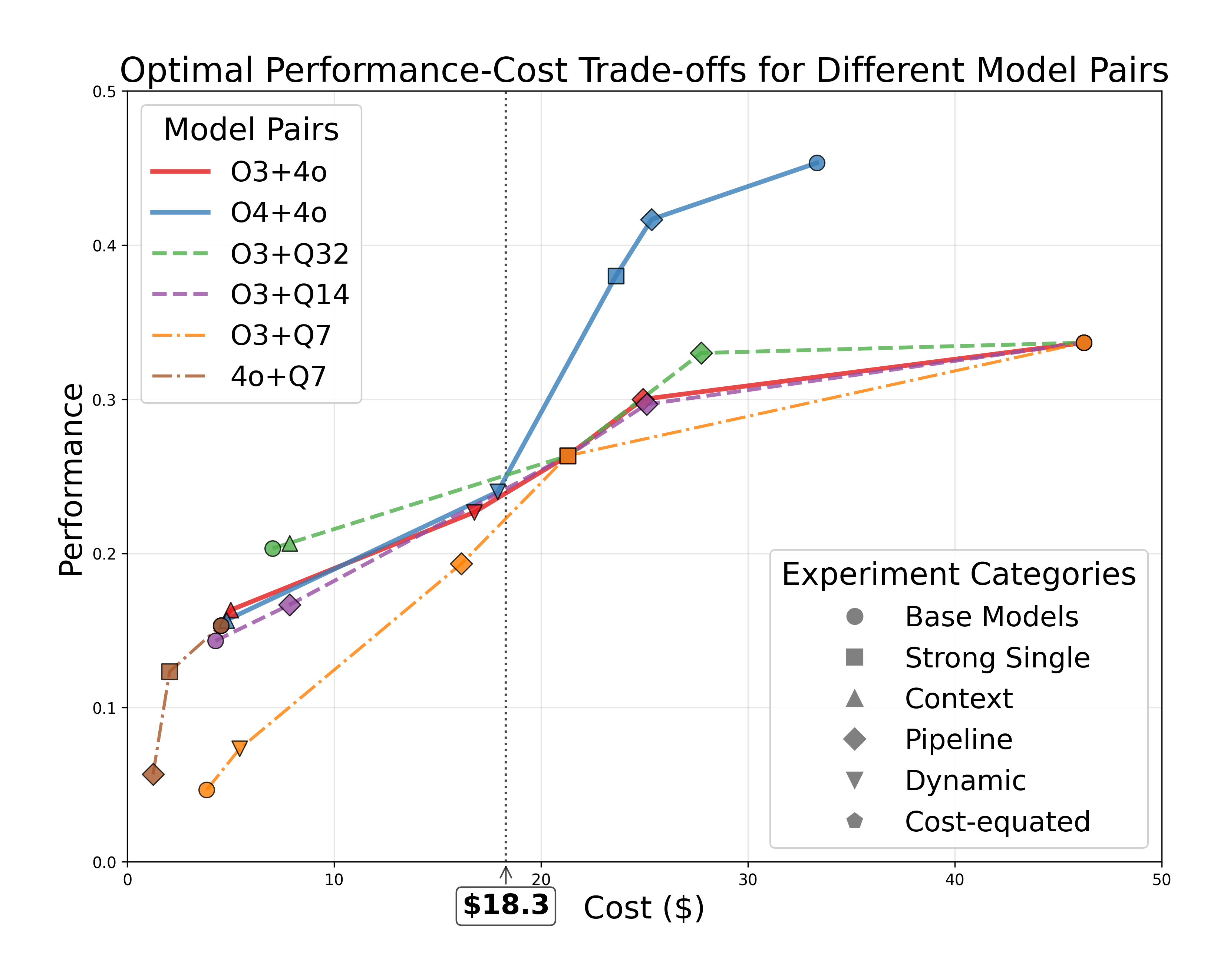}
    \caption{Performance vs. Cost curves for different Strong-Weak Model Pairs. O3 - O3-mini; O4 - O4-mini; 4o - GPT-4o-mini; Qx - Qwen2.5-Coder-xB. Detailed pairwise results in Appendix \autoref{fig:perf_cost_comparison} and Appendix Tables \ref{tab:results-o4-mini-gpt-4o-mini}-\ref{tab:results-gpt-4o-mini-qwen25coder7b}.}
    \label{fig:combined_perf_vs_cost}
\end{figure}

\section{Main Results}
\label{sec:results_and_analysis}

We focus our evaluation on the generator, as \texttt{voyage-code-3} achieves strong retrieval ($\sim$92\% recall, $\sim$76\% MRR).
Our findings uncover consistent patterns when strong–weak collaboration succeeds or fails (Appendix \autoref{fig:perf_cost_comparison},  Appendix Tables \ref{tab:results-o4-mini-gpt-4o-mini}-\ref{tab:results-gpt-4o-mini-qwen25coder7b}.

\start{Cost-equated baselines vs. Strong-Weak collaboration}
As shown in \autoref{fig:combined_perf_vs_cost}, Appendix \autoref{fig:perf_cost_comparison} and Appendix Tables \ref{tab:results-o4-mini-gpt-4o-mini}-\ref{tab:localization_gpt-4o-mini_qwen7b} and Appendix \autoref{fig:perf_cost_comparison}, across nearly all model pairs, the cost-equated baselines, i.e. sampling multiple times to match the strong LLM's cost, underperform compared to collaboration. Surprisingly, some baselines even hurt weak LLM's performance, which is likely due to faltering patch selection.
Our best collaboration strategy, Strong LM First, achieves a 0.4167 resolution rate, which is $\sim$92\% more than the best corresponding cost-equated baseline, i.e., best-of-n for GPT-4o-mini cost-equated with O4-mini. 

\start{Efficacy of Methods Across Taxonomy Groups}
Analyzing by taxonomy (\S\ref{sec:approach}), on average, pipeline and context-based strategies yield the highest cost-efficiency across model pairs, followed by dynamic methods and \textit{Best of n}, with self-consistency approaches being the least efficient (Appendix \S\ref{sec:stats}).
However, average efficiency alone is not a reliable indicator of the best method. Performance–cost curves (Figure~\ref{fig:combined_perf_vs_cost}) often intersect, showing that \underline{Weak LM First} is optimal when budget is low due to minimal strong LM usage, while \underline{Strong LM First} dominates once budget increases. For example, with O3-mini + Qwen2.5-Coder-32B, \underline{Strong LM First} achieves equivalent performance to the strong model at just $\sim$60\% cost. We also observe a regime shift around the \$18.3 mark as Figure \ref{fig:combined_perf_vs_cost} shows, below which O3 + Qwen2.5-Coder-32B performs best, whereas above it, O4 + GPT-4o-mini becomes more cost-effective.

These results suggest that no method is universally optimal. Instead, choosing the best strategy depends on the deployment scenario, specifically the available budget and required performance, which is enabled by plots like Figure \ref{fig:combined_perf_vs_cost}. Given a minimum performance threshold and a budget cap, the optimal method corresponds to the curve that reaches the highest resolution rate within the feasible region i.e., the top-left quadrant defined by those constraints. For instance, if the target is at least 20\% resolution within a \$20 budget, \textit{Weak Router} with O4-mini + GPT-4o-mini emerges as the most cost-effective choice.
These observations yield actionable guidance:
(1) Cost-equated weak-only methods are inefficient;
(2) Pipeline-based methods outperform context-only methods when budget allows;
(3) \textit{Weak LM First} and \textit{Weak Router} are strong choices under tight budgets; and
(4) \textit{Strong LM First} performs best in higher-budget regimes, often approaching strong LM performance at reduced cost.

\subsection{Interesting cases}

\start{Instance-level help is better than repo-level help}
Repo-level context (e.g., summaries, structure, QA pairs) consistently failed to improve weak LM performance. Such information is too coarse and may distract from instance-specific signals. Even few-shot examples, whether random or similar, often reduced accuracy. In contrast, instance-level augmentation (plans, QA pairs) significantly boosted resolution rates, justifying their higher cost.

\start{Weak Router is better than Strong Router}
Counterintuitively, \textit{Weak Router} frequently outperformed \textit{Strong Router}, both in accuracy and cost. For example, with O4-mini + Qwen2,5-Coder-32B, \textit{Weak Router} achieved 6 percentage points higher resolution at $\sim$20\% lower cost. We suspect that stronger models
, while good at problem solving, may “overthink” routing decisions~\cite{cuadron2025dangeroverthinkingexaminingreasoningaction}. Only with very weak LMs (e.g., Qwen-7B) did strong routing slightly edge ahead, suggesting weak routing is often the more efficient choice.

\start{Prompt Reduction trades turn-wise validity for overall performance}
\textit{Prompt Reduction} offers a surprising trade-off where it lowers the valid patch rate to $\sim$65\%, which is well below the $\sim$95\% average of other methods, yet often achieves higher resolution rates overall. Due to aggressive pruning of irrelevant context by the weak LM, the strong LM is forced to focus on the most salient code, leading to more successful fixes despite more retries. In effect, Prompt Reduction sacrifices turn-wise reliability in favor of sharper attention, making it a high-variance but high-upside strategy, especially when correctness matters more than efficiency.

\section{Conclusion}
\label{sec:conclusion}

We present a detailed taxonomy of strong-weak LLM collaboration techniques for repository-level code generation.
Our results show substantial cost-efficiency gains with pipeline and context-based collaboration, maximizing average efficiency. 
However, no method is universally optimal, and the ideal choice depends on budget and performance constraints. 
We offer actionable principles for selecting when to prioritize weak models, apply dynamic routing, or escalate to hybrid pipelines.
Our insights will help practitioners deploy accurate LLM systems aligned with realistic efficiency constraints.
Future work can extend this taxonomy to more complex approaches and domains for improved cost-performance tradeoffs.
\section{Limitations and Potential Risk}

For simplicity of experimentation, we restrict the scope of this study to Agentless Lite, a simple RAG + Code generation framework. However, we believe that the methods we consider are generalizable and simple enough to be extended to more complex architectures. We also limit the scope of this study to the code generation domain, however, parallels can easily be drawn to other domains and the methods can be tweaked to suit those applications. To curb costs for experimentation, we use the Lite subset of the SWE-Bench dataset. We focus the study only on inference-based approaches and do not consider any training or finetuning-related approaches since those often require expensive hardware setups which defeats the overall purpose of cost efficiency. We measure cost via Token / API usage and do not consider factors like latency or energy consumption since those would vary significantly with hardware. 
Potential risks of our work are similar to risks largely associated with LLM-based code generation in general. These include but are not limited to the methods studied being used to purposefully or inadvertently generate vulnerable or malicious code.

\section{Scientific Artifacts}

We use a variety of proprietary and open sourced LLMs for our study. We strictly use commercial APIs to access proprietary LLMs following the conditions outlined by the respective companies. We use an open source framework, Agentless Lite, for our experiments. We use the test set of the SWE-Bench Lite Benchmark for our experiments. To the best of our knowledge, it does not contains personally identifying information or offensive content.

\bibliography{custom}

\clearpage


\appendix

\begin{figure*}[!h]
    \centering
    \includegraphics[width=0.7\linewidth]{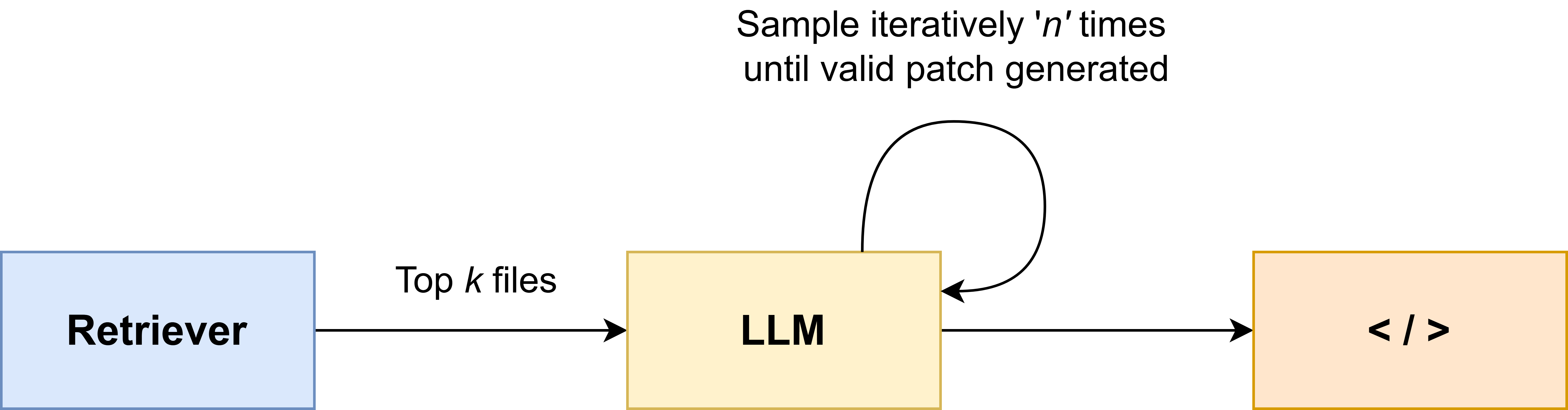}
    \caption{The Agentless Lite Framework: RAG + Code Generation}
    \label{fig:agentless_lite}
\end{figure*}

\section{Appendix}
\label{sec:appendix}


\subsection{Detailed Related Work}

\start{Repository-Level Code Generation}
Repository-Level Code Generation is a complex problem that requires LLMs to leverage the context of an entire repository and understand the structure and interleaving dependencies. The SWE-Bench benchmark \cite{jimenez2024swebench} exemplifies this in the form of issue-resolution, a real-world repository-level code generation task, where the system must make edits to the code given an issue description and the relevant repository snapshot. This task can mainly be broken down into two sub-tasks: Fault Localization and Program Repair. Recent methods often use retrieval or structured pipelines to tackle this problem. For example, Retrieval Augmented Code Generation (RACG) frameworks first retrieve relevant code or documentation from the repository to aid in efficient fault localization before generation \cite{wang-etal-2025-coderag, li2025coderagsupportivecoderetrieval}. While there exist agentic techniques like SWE-Agent \cite{yang2024sweagentagentcomputerinterfacesenable} that are able to achieve a significant performance on this task, it often comes at a steep cost involving multiple calls to expensive LLMs. On the other hand, Agentless \cite{agentless} entirely avoids multi-turn agent calls and instead sticks to a fixed three-step process involving localization, repair and validation, showing strong performance and low cost on SWE-Bench.

\subsection{Hyperparameters and Implementation Details}\label{sec:hyperparameters}

We plan on open-sourcing our code soon. We set the \texttt{max\_files} parameter for Agentless Lite to 5.
For OpenAI's thinking models like O3-mini and O4-mini, we set the \texttt{reasoning\_effort} parameter to \texttt{high}. We start the generation at a base temperature of \texttt{0.0}, which is gradually increased by \texttt{0.1} every time the model fails to generate a valid patch. We set the max number of retries as 10 for the base settings. 
All open-source model inferences were run using \texttt{vLLM} on 4 L40 GPUs, leveraging tensor parallelism for efficient distributed execution. We configured the server with \texttt{--tensor-parallel-size 4} and set \texttt{--gpu-memory-utilization 0.8} to optimize memory usage without overcommitment. Inference was performed in \texttt{bfloat16} precision with \texttt{--enforce-eager} execution to reduce kernel launch latency. To support long-context models, we enabled \texttt{chunked prefill} and set the \texttt{rope\_scaling} parameter to use \texttt{YARN} with a factor of 4.0, extending support to sequences up to 32k tokens. Additional parameters included a maximum of 32,768 batched tokens, 8 concurrent sequences, and 16GB of CPU swap space to handle occasional memory spikes.

\start{Cost-equated weak LM}
We first study several baselines that \emph{only} involve the weak LM~\cite{mcdonald-etal-2025-afford}.
Self-consistency \citep{wang2023selfconsistencyimproveschainthought} enhances weak LMs by sampling $n$ diverse outputs, where $n$ $\approx$ Cost$_{strong}$ / Cost$_{weak}$ and selecting the most consistent one via:
\circled{1} \underline{Majority Voting ($SC_m$)}, where we select the most frequent patch,
\circled{2} \underline{Clustering ($SC_c$)}, where we cluster the $n$ candidate patches and select a patch at random from the largest cluster. We cluster the patches based on similarity calculated using \texttt{difflib.SequenceMatcher} 
\circled{3} \underline{Universal ($SC_u$)}, where we follow \citet{chen2023universalselfconsistencylargelanguage} and prompt the weak LM itself to select the final patch.
We also experiment on a ``cheating'' method, \circled{4} \underline{Best of n}, which checks whether any of the candidate patches is correct and serves as an upper bound for candidate selection methods.

\start{Evaluation Metrics}
The following metrics are reported for each method: (1) \textit{Resolution Rate}: Proportion of instances for which the generated patch successfully resolves the issue;
(2) \textit{Average \#Iterations}: The average number of calls made to strong and weak models, denoted by \textit{s} and \textit{w} respectively;
(3) \textit{Valid Patch Rate}: Fraction of instances for which the system was able to generate a patch that adheres to the required format;
(4) \textit{Cost}: Total generation cost (\$) 
to process all 300 SWEBench-Lite instances,
including additional method costs, but excluding retrieval cost, which remains constant across all methods, 
and
(5) \textit{Efficiency Score}: The ratio of Resolution Rate to Cost.

\subsection{Prompts}\label{sec:prompts}

\begin{table*}[!h]
\centering
\begin{threeparttable}
\resizebox{\textwidth}{!}{%
\begin{tabular}{@{}lccccc@{}}
\toprule
\textbf{Experiment} & \textbf{Resolution Rate} & \textbf{Avg. \#iterations} & \textbf{Valid Patch Rate} & \textbf{Total Generation Cost} & \textbf{Efficiency} \\ \midrule
o3-mini\tnote{1}             & 0.3367 & 1.83 & 0.977 & 46.224     & 0.0073 \\
o4-mini\tnote{2}             & \textbf{0.4533} & \textbf{1.32} & \textbf{0.993} & 33.328     & 0.0136 \\
claude-3.5-sonnet\tnote{3}   & 0.2667 & 2.13 & 0.957 & \textbf{140.126} & 0.0019 \\
gpt-4o-mini\tnote{4}         & 0.1533 & 2.09 & 0.960 & 4.529      & \textbf{0.0338} \\
qwen2.5-coder-7b\tnote{5}    & 0.0467 & 6.48 & 0.617 & 3.839*     & 0.0122 \\
qwen2.5-coder-14b            & 0.1433 & 2.91 & 0.937 & 4.257*     & 0.0337 \\
qwen2.5-coder-32b            & 0.2033 & 2.35 & 0.967 & 7.021*     & 0.0290 \\
r1-distill-qwen-7b\tnote{6}  & 0.0033 & 9.91 & 0.013 & 115.161*   & 0.0000 \\ \bottomrule
\end{tabular}%
}
\begin{tablenotes}
\footnotesize
\item[1] \texttt{o3-mini-2025-01-31} from \url{https://platform.openai.com/docs/models/o3-mini}
\item[2] \texttt{o4-mini-2025-04-16} from \url{https://platform.openai.com/docs/models/o4-mini}
\item[3] \texttt{claude-3-5-sonnet-20241022} from \url{https://docs.anthropic.com/en/docs/about-claude/models/all-models}
\item[4] \texttt{gpt-4o-mini-2024-07-18} from \url{https://platform.openai.com/docs/models/gpt-4o-mini}
\item[5] Qwen-2.5-Coder-Instruct series models from \url{https://huggingface.co/Qwen}
\item[6] \texttt{DeepSeek-R1-Distill-Qwen-7B} from \url{https://huggingface.co/deepseek-ai/DeepSeek-R1-Distill-Qwen-7B}
\end{tablenotes}
\end{threeparttable}
\caption{Base performance of various models with Agentless Lite. Efficiency is calculated as resolution rate divided by generation cost. Models like \texttt{claude-3-5-sonnet-20241022} and \texttt{r1-distill-qwen-7b} are excluded from our main study due to poor cost-effectiveness. * - Estimated cost based on comparable API pricing.}
\label{tab:base_performance}
\end{table*}

\subsubsection{Agentless Lite Base Prompt}
\begin{lstlisting}[style=promptstyle]
We are currently solving the following issue within our repository. Here is the issue text:
--- BEGIN ISSUE ---
{problem_statement}
--- END ISSUE ---

Below are some code segments, each from a relevant file. One or more of these files may contain bugs:
--- BEGIN FILE ---
{retrieval}
--- END FILE ---

Please first localize the bug based on the issue statement, and then generate *SEARCH/REPLACE* edits to fix the issue.

Every *SEARCH/REPLACE* edit must use this format:
1. The file path
2. The start of search block: <<<<<<< SEARCH
3. A contiguous chunk of lines to search for in the existing source code
4. The dividing line: =======
5. The lines to replace into the source code
6. The end of the replace block: >>>>>>> REPLACE

Here is an example:

```python
### mathweb/flask/app.py
<<<<<<< SEARCH
from flask import Flask
=======
import math
from flask import Flask
>>>>>>> REPLACE
```

Please note that the *SEARCH/REPLACE* edit REQUIRES PROPER INDENTATION. If you would like to add the line '        print(x)', you must fully write that out, with all those spaces before the code!
Wrap the *SEARCH/REPLACE* edit in blocks ```python...```.
\end{lstlisting}


\subsubsection{Repo Summary}
\begin{lstlisting}[style=promptstyle]
I need you to provide high-level insights about the following repository: {repo_name}

Based on the repository structure and README below, generate a comprehensive overview of this repository that could help guide a language model in solving technical issues.

Repository Structure:
{repo_structure}

README Content:
{readme_content}

Please provide the following insights. For each point, provide concrete details and specific examples from the codebase - high-level doesn't mean vague, it means providing a clear architectural overview with specific names, patterns, and implementations:

1. Core Purpose and Functionality: 
    - What specific problem does this repository solve?
    - What are its primary features and capabilities?

2. Main Architectural Patterns:
    - Identify concrete architectural patterns used in this codebase
    - EXAMPLE: Plugin based architecture, layered architecture, etc

3. Module Organization:
    - Name the specific key modules and their exact responsibilities
    - EXAMPLE: I/O module, error-handling module, etc

4. Key Abstractions and Concepts:
    - List the actual fundamental abstractions used in the codebase
    - EXAMPLE: Quantity class for numerical values, Logger class for logging, etc

5. Design Patterns:
    - Identify specific recurring code patterns with examples
    - EXAMPLE: Factory methods, Decorators, etc

6. Error Handling Approaches:
    - Describe precise error handling mechanisms used in the codebase
    - EXAMPLE: Custom exception hierarchies, warnings, etc

Focus on providing actionable architectural insights that would be valuable for understanding the repository's design philosophy and core abstractions. Your response should contain specific implementation details that would help someone understand how to navigate, extend, and debug the codebase to solve issues.
\end{lstlisting}


\subsubsection{Repo Level QA Pairs}
\begin{lstlisting}[style=promptstyle]
I need you to generate a comprehensive FAQ about the repository: {repo_name}

Based on the repository structure and README below, create a detailed set of technical FAQs that would help a developer solve issues in this codebase. These FAQs should serve as guidance for someone who is trying to resolve bugs or implement new features.

Repository Structure:
{repo_structure}

README Content:
{readme_content}

Please generate 15-20 frequently asked questions with detailed answers about:

1. Code Organization and Architecture:
   - How is the codebase structured?
   - What are the key modules and their responsibilities?
   - How do the different components interact?

2. Common Patterns and Conventions:
   - What design patterns are commonly used?
   - What are the naming conventions and code style expectations?
   - Are there specific patterns for implementing new features?

3. Typical Debugging Approaches:
   - What are common error patterns and their solutions?
   - How to debug specific types of issues in this codebase?
   - What are common pitfalls when modifying this code?

4. Implementation Details:
   - How are core abstractions implemented?
   - What are the key algorithms or data structures used?
   - How does the error handling system work?

5. Testing Considerations:
   - How is testing typically done in this codebase?
   - What should be considered when writing tests?
   - Are there common test fixtures or utilities?

For each question, provide detailed, specific answers with concrete examples from the codebase when possible. Focus on information that would be most valuable to someone trying to fix bugs or implement new features. The FAQs should reflect the actual patterns and practices used in this specific repository, not generic software development advice.
\end{lstlisting}


\subsubsection{Repo Structure}
\begin{lstlisting}[style=promptstyle]
We are currently solving the following issue within our repository. Here is the issue text:
--- BEGIN ISSUE ---
{problem_statement}
--- END ISSUE ---

Below are some code segments, each from a relevant file. One or more of these files may contain bugs:
--- BEGIN FILE ---
{retrieval}
--- END FILE ---

To help you better understand the contexts of the code segments, we provide a set of dependencies of the code segments. 
The dependencies reflect how the functions/classes in the code segments are referenced in the codebase. 

--- BEGIN DEPENDENCIES ---
{dependencies}
--- END DEPENDENCIES ---

Please first localize the bug based on the issue statement, and then generate *SEARCH/REPLACE* edits to fix the issue.

Every *SEARCH/REPLACE* edit must use this format:
1. The file path
2. The start of search block: <<<<<<< SEARCH
3. A contiguous chunk of lines to search for in the existing source code
4. The dividing line: =======
5. The lines to replace into the source code
6. The end of the replace block: >>>>>>> REPLACE

Here is an example:

```python
### mathweb/flask/app.py
<<<<<<< SEARCH
from flask import Flask
=======
import math
from flask import Flask
>>>>>>> REPLACE
```

Please note that the *SEARCH/REPLACE* edit REQUIRES PROPER INDENTATION. If you would like to add the line '        print(x)', you must fully write that out, with all those spaces before the code!
Wrap the *SEARCH/REPLACE* edit in blocks ```python...```.
\end{lstlisting}


\subsubsection{Few Shot Examples}
\begin{lstlisting}[style=promptstyle]
Here are some {similar}example issues from the same repository along with the target file that were changed and final patch generated by an expert{successful}:
--- BEGIN EXAMPLES ---
{few_shot_examples}
--- END EXAMPLES ---

We are currently solving the following issue within our repository. Here is the issue text:
--- BEGIN ISSUE ---
{problem_statement}
--- END ISSUE ---

Below are some code segments, each from a relevant file. One or more of these files may contain bugs:
--- BEGIN FILE ---
{retrieval}
--- END FILE ---

Please first localize the bug based on the issue statement, and then generate *SEARCH/REPLACE* edits to fix the issue.

Every *SEARCH/REPLACE* edit must use this format:
1. The file path
2. The start of search block: <<<<<<< SEARCH
3. A contiguous chunk of lines to search for in the existing source code
4. The dividing line: =======
5. The lines to replace into the source code
6. The end of the replace block: >>>>>>> REPLACE

Here is an example:

```python
### mathweb/flask/app.py
<<<<<<< SEARCH
from flask import Flask
=======
import math
from flask import Flask
>>>>>>> REPLACE
```

Please note that the *SEARCH/REPLACE* edit REQUIRES PROPER INDENTATION. If you would like to add the line '        print(x)', you must fully write that out, with all those spaces before the code!
Wrap the *SEARCH/REPLACE* edit in blocks ```python...```.
\end{lstlisting}


\subsubsection{Plan}
\begin{lstlisting}[style=promptstyle]
We are currently solving the following issue within our repository. Here is the issue text:
--- BEGIN ISSUE ---
{problem_statement}
--- END ISSUE ---

Below are some code segments, each from a relevant file. One or more of these files may contain bugs:
--- BEGIN FILE ---
{retrieval}
--- END FILE ---

Please analyze the issue and provide a detailed plan to fix it. Do NOT generate any code patches or specific edits.

Your plan should include:

1. Bug localization: Identify which file(s) contain the bug based on the issue statement
2. Root cause analysis: Explain why the bug is occurring
3. Solution approach: Describe conceptually how to fix the issue
4. Implementation strategy: Outline the logical steps needed to implement the solution

Keep your analysis focused on the problem-solving approach rather than specific code changes.
\end{lstlisting}


\subsubsection{Instance Level QA Pairs}
\begin{lstlisting}[style=promptstyle]
We are currently solving the following issue within our repository. Here is the issue text:
--- BEGIN ISSUE ---
{problem_statement}
--- END ISSUE ---

Below are some code segments, each from a relevant file. One or more of these files may contain bugs:
--- BEGIN FILE ---
{retrieval}
--- END FILE ---

Please create a detailed FAQ (Frequently Asked Questions) document that would help a junior developer understand and fix this specific issue. Do NOT generate any code patches or specific edits.

Your FAQ should include 7-10 questions and answers about:

1. Issue Understanding:
   - What is the exact problem described in the issue?
   - What are the expected vs. actual behaviors?
   - What conditions trigger this issue?

2. Codebase Navigation:
   - Which specific files and functions are most relevant to this issue?
   - What are the key components involved in this functionality?
   - How do these components interact?

3. Technical Analysis:
   - What are the potential root causes of this issue?
   - What code patterns or anti-patterns might be contributing to the bug?
   - What specific edge cases might not be handled correctly?

4. Implementation Guidance:
   - What approaches could be used to fix this issue?
   - What implementation pitfalls should be avoided?
   - How should the solution be tested?

5. Codebase Specifics:
   - What patterns or conventions does this codebase use that are relevant to the fix?
   - What existing helper functions or utilities could be leveraged?
   - What dependencies or side effects need to be considered?

Make your questions and answers detailed, specific to this issue, and include concrete references to the code when possible. Avoid generic programming advice - focus on information that directly helps solve this specific issue.
\end{lstlisting}


\subsubsection{Prompt Reduction}
\begin{lstlisting}[style=promptstyle]
We are currently solving the following issue within our repository. Here is the issue text:
--- BEGIN ISSUE ---
{problem_statement}
--- END ISSUE ---

Below are some code segments, each from a relevant file. One or more of these files may contain bugs:
--- BEGIN FILE ---
{retrieval}
--- END FILE ---

Your task is to identify the most relevant code sections that contain the bug or need to be modified to fix the issue.
Please output only the file paths and relevant code sections in this format:

### <file_path>
```python
# relevant code section 1
```

### <file_path>
```python
# relevant code section 2
```

Only include the minimum necessary code with sufficient context to understand and fix the issue. Ensure that the code is complete and valid Python code.
Don't include any explanations or reasoning - just the file paths and code sections.
\end{lstlisting}


\subsubsection{Weak / Strong Router}
\begin{lstlisting}[style=promptstyle]
You are an expert software engineer tasked with analyzing software issues to determine the most efficient debugging approach.

Please analyze the following issue and codebase:

Issue description:
--- BEGIN ISSUE ---
{problem_statement}
--- END ISSUE ---

Relevant code:
--- BEGIN CODE ---
{retrieval}
--- END CODE ---

Your goal is to classify this issue as either SIMPLE or COMPLEX.

SIMPLE issues typically have these characteristics:
- Clear localization of the bug in the code
- Straightforward cause-effect relationship
- Relatively isolated impact (limited to one function or module)
- Common programming patterns or errors
- Solution likely follows established best practices

COMPLEX issues typically have these characteristics:
- Multiple components involved in the bug
- Subtle interactions between different parts of the codebase
- Requires deep reasoning about codebase architecture
- Edge cases that are difficult to identify
- May require creative or non-obvious solutions

Based solely on your analysis of the issue and code, respond with ONLY one of these two options:
- SIMPLE
- COMPLEX

Provide no explanation, reasoning, or additional text - just output the single classification word.
\end{lstlisting}


\subsubsection{Universal Self-Consistency Patch Selection}
\begin{lstlisting}[style=promptstyle]
I have generated the following {n_samples} potential solutions to fix this issue:

# Problem Statement
{problem_statement}

# Relevant Files
{formatted_files}

# Generated Solutions
{patches}

Based on the above solutions, select the most consistent and correct solution. Analyze the similarities and differences between the solutions, and select the one that best addresses the problem statement while making minimal and precise changes. 

Your selection should be based on:
1. Correctness (does it solve the issue described in the problem statement)
2. Consensus (do multiple solutions agree on a similar approach)
3. Simplicity (does it make minimal necessary changes)

Return your selection as "SELECTED_PATCH: X" where X is the number of the chosen patch (1 to {n_samples}) and then explain your reasoning.
\end{lstlisting}
\begin{table*}[!ht]
\centering
\begin{tabular}{@{}llr@{}}
\toprule
\textbf{Level} & \textbf{Group} & \textbf{Least Sq Mean} \\
\midrule
Repo Structure                                & A         & 0.02818333 \\
Weak LM First                                 & A B       & 0.02365000 \\
Repo Level QA Pairs                           & A B C     & 0.02128333 \\
Strong LM Single Attempt                      & A B C     & 0.02100000 \\
Few Shot Examples (1 Shot Random)            & A B C D   & 0.02028333 \\
Few Shot Examples (1 Shot Similarity)        & A B C D E & 0.01988333 \\
Repo Summary                                  & B C D E   & 0.01821667 \\
Strong LM First                               & B C D E F & 0.01640000 \\
Few Shot Examples (5 Shot Similarity)        & C D E F G & 0.01501667 \\
Few Shot Examples (5 Shot Random)            & C D E F G & 0.01483333 \\
Weak Router                                   & D E F G H & 0.01170000 \\
Prompt Reduction                              & E F G H   & 0.01135000 \\
Plan                                          & F G H     & 0.00801667 \\
Instance Level QA Pairs                       & G H       & 0.00750000 \\
Best of N                                     & H         & 0.00605000 \\
Strong Router                                 & H         & 0.00585000 \\
Self-Consistency – Clustering                 & H         & 0.00491667 \\
Self-Consistency – Majority Vote              & H         & 0.00486667 \\
Self-Consistency – Universal                  & H         & 0.00475000 \\
\bottomrule
\end{tabular}
\caption{Least Squares Mean Differences for Efficiency across Methods. Methods sharing a letter are not significantly different (Student’s t, $\alpha=0.05$, $t=1.98667$).}
\label{tab:lsmeans_efficiency}
\end{table*}

\subsection{Statistical Analysis}\label{sec:stats}

In order to systematically compare average efficiency, accuracy, and cost for 19 separate method configurations under 5 different method groups, we evaluated performance on SWE-Bench Lite over 6 strong-weak model pairs using 2 different ANOVA models for each of the 3 dependent measures, namely accuracy, efficiency, and cost.

For the first set of ANOVAs, we include Model Pair and Method Group as independent variables so that we can capture average trends while controlling for the effect of model pair.  Both variables were highly significant in all three models (p < .00001), and each of the models captured between 66\% and 80\% of the variance in dependent measure, thus indicating that these models tell a meaningful story about what affects these dependent measures.   For accuracy, Pipeline approaches were significantly better than all others.  Self-Consistency and Context were significantly worse than all others.  Best of n and Dynamic were in the middle.  For cost, Self-Consistency and Best of n were significantly more than all others.  Context was significantly cheaper than all others.  And Pipeline and Dynamic were less than Self-Consistency and Best of n and more than Context.  In terms of average efficiency, Pipeline and Context were significantly better than all others.

Taking a look at a higher level of granularity, we constructed a second set of ANOVAs shown in Appendix \autoref{tab:lsmeans_efficiency} with the specific methods rather than method groups.  In this case, we see some overlap between groups of methods in terms of their accuracy, cost, and efficiency, but the general trends regarding the sets of methods are still very visible.  Again, the two independent variables were highly significant in each model, and the amount of variance in dependent measure captured by the models ranged between 59\% and 87\%.  For performance, \underline{Strong LM First} and \underline{Strong LM Single Attempt} were significantly higher than all others.  In terms of cost, there were three separable groups, where the three Self-Consistency approaches, \underline{Prompt Reduction}, and \underline{Strong Router} were significantly higher than all others, followed by \underline{Plan}, \underline{Instance level QA Pairs}, \underline{Strong LM First}, \underline{Strong LM Single Attempt}, and \underline{Weak Router} were significantly less expensive than those but significantly more expensive than the remaining methods.  In terms of efficiency, \underline{Repo Structure} is the best, but it is not significantly better than \underline{Weak LM First}, \underline{Repo-Level QA Pairs}, \underline{Strong LM Single Attempt}, and \underline{Few shot (1 shot random and 1 shot similarity)}.



\begin{figure*}[!h]
    \centering
    \begin{tabular}{ccc}
        \begin{subfigure}[b]{0.32\textwidth}
            \centering
            \includegraphics[width=\textwidth]{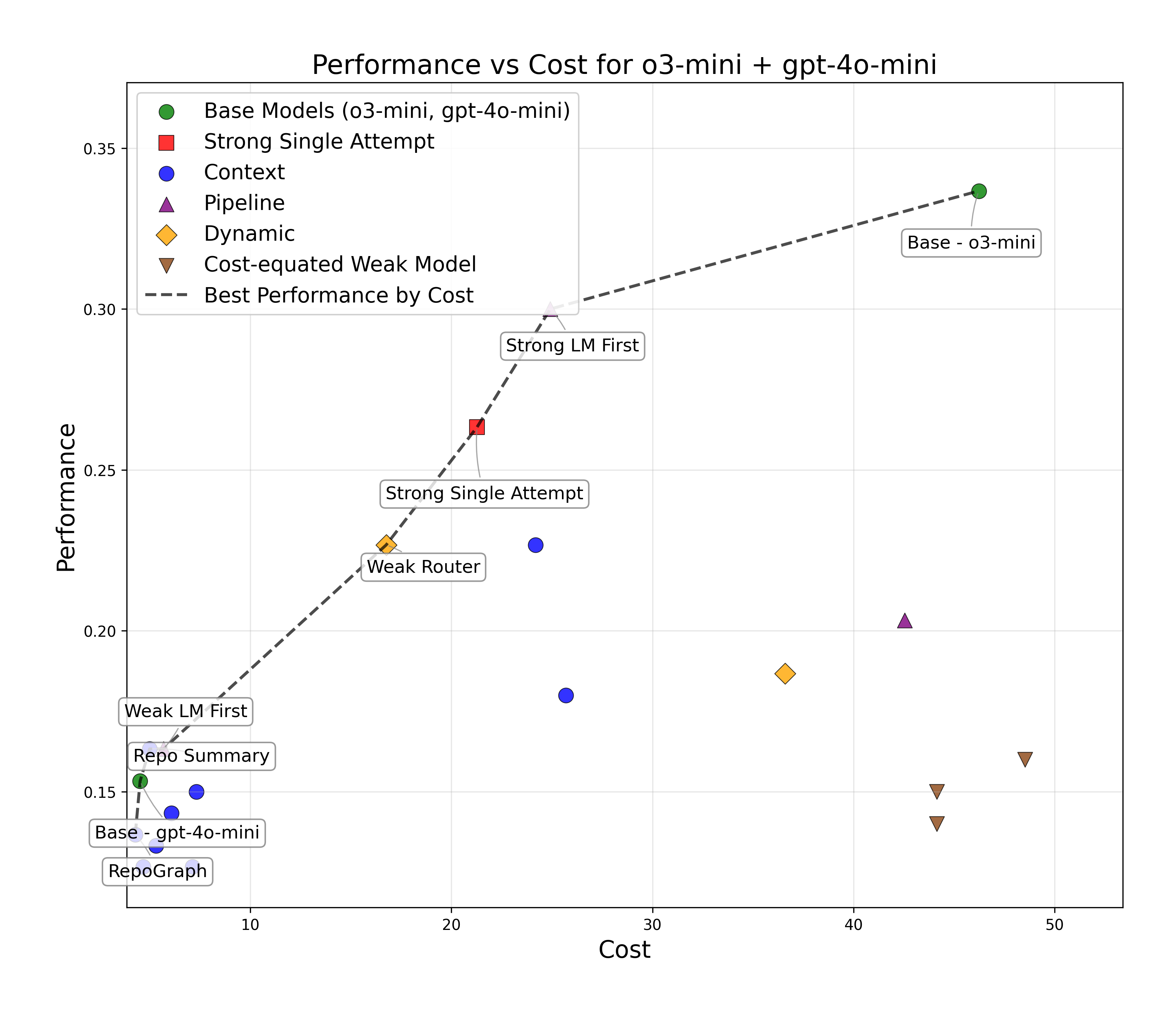}
            \caption{\centering O3-mini (0.3367) + \newline GPT-4o-mini (0.1533)}
            \label{fig:subfig1}
        \end{subfigure} &
        \begin{subfigure}[b]{0.32\textwidth}
            \centering
            \includegraphics[width=\textwidth]{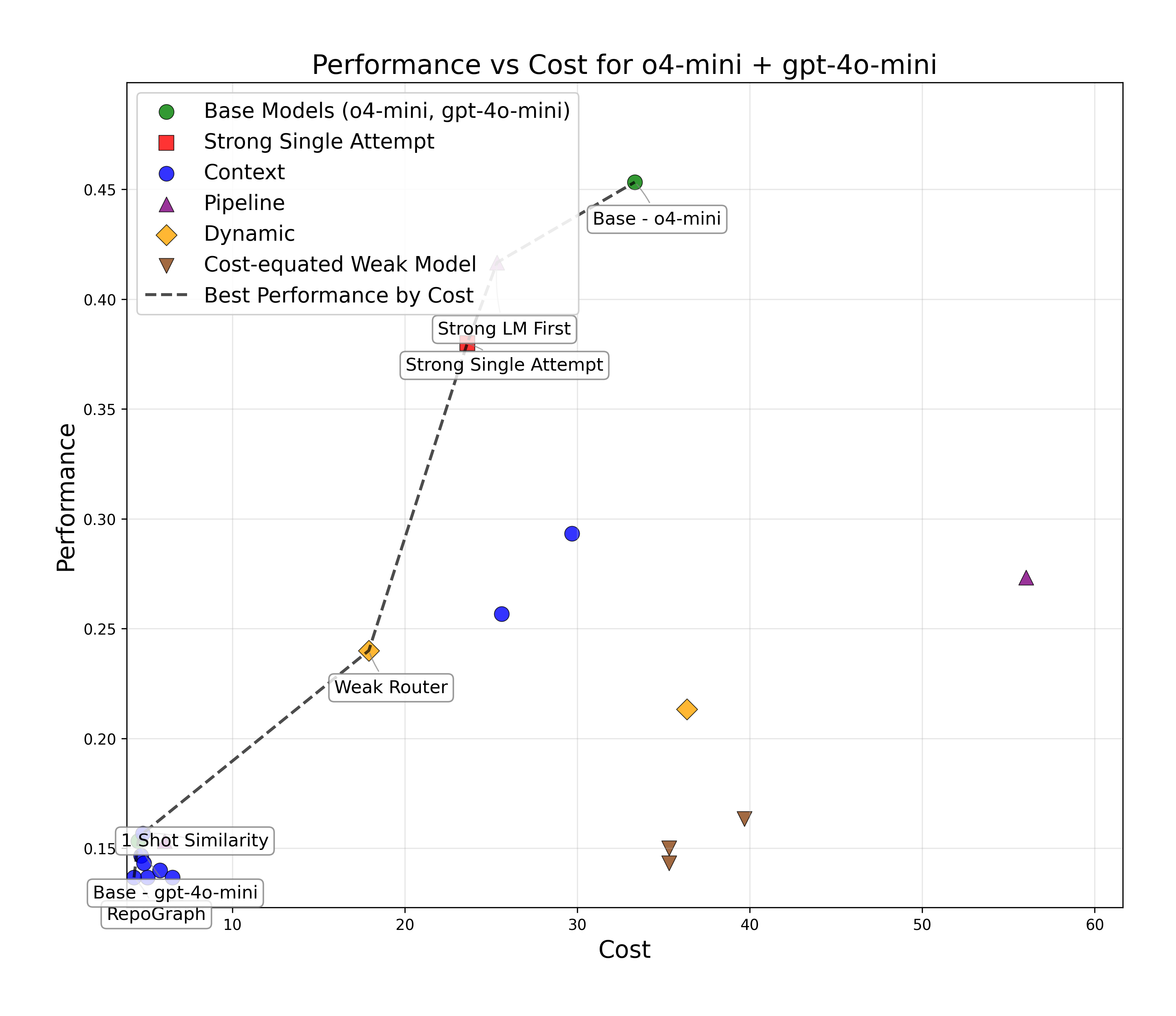}
            \caption{\centering O4-mini (0.4533) + \newline GPT-4o-mini (0.1533)}
            \label{fig:subfig2}
        \end{subfigure} &
        \begin{subfigure}[b]{0.32\textwidth}
            \centering
            \includegraphics[width=\textwidth]{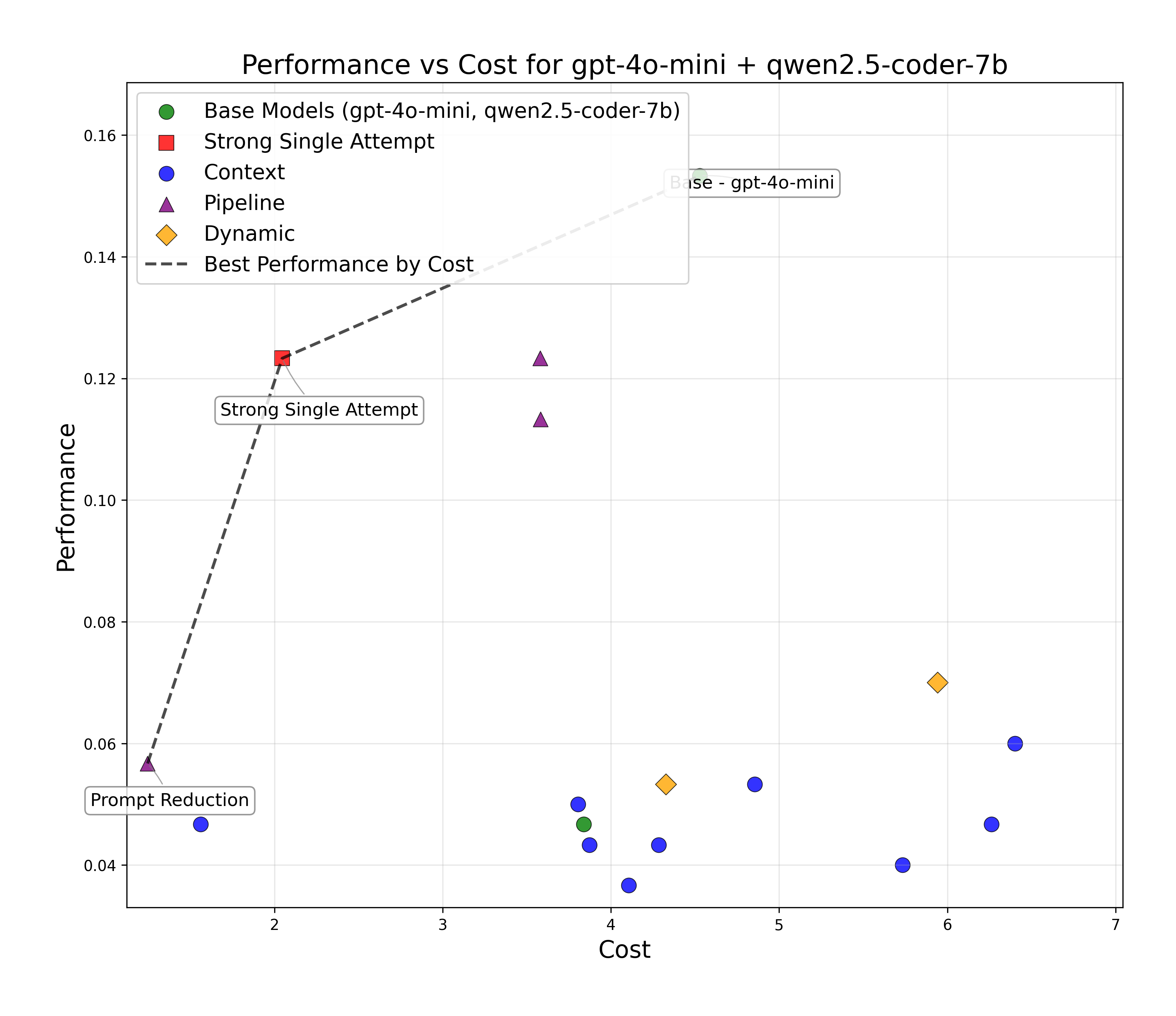}
            \caption{\centering GPT-4o-mini (0.1533) + \newline Qwen2.5-Coder-7B (0.0467)}
            \label{fig:subfig3}
        \end{subfigure} \\
        \begin{subfigure}[b]{0.32\textwidth}
            \centering
            \includegraphics[width=\textwidth]{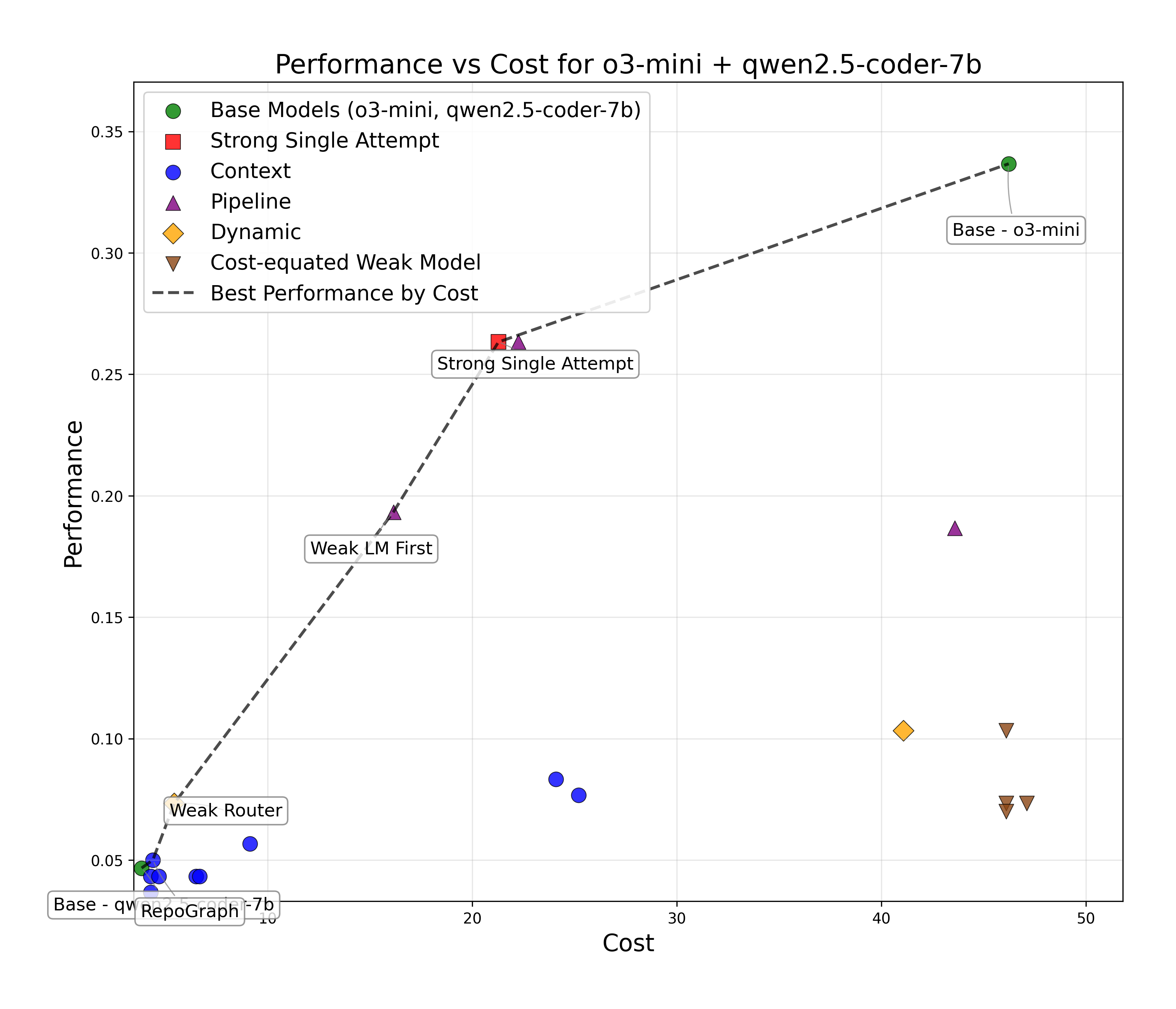}
            \caption{\centering O3-mini (0.3367) + \newline Qwen2.5-Coder-7B (0.0467)}
            \label{fig:subfig4}
        \end{subfigure} &
        \begin{subfigure}[b]{0.32\textwidth}
            \centering
            \includegraphics[width=\textwidth]{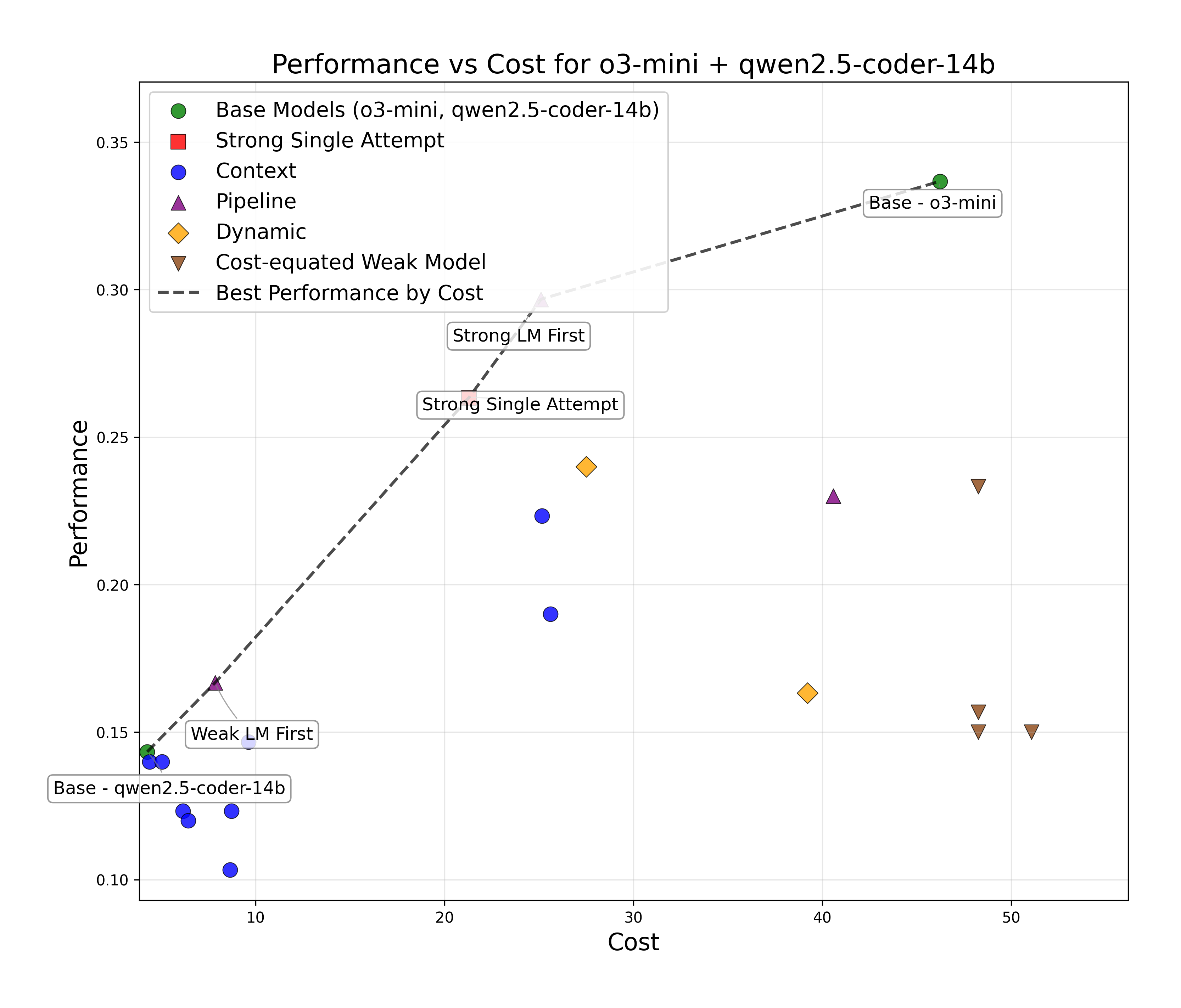}
            \caption{\centering O3-mini (0.3367) + \newline Qwen2.5-Coder-14B (0.1433)}
            \label{fig:subfig5}
        \end{subfigure} &
        \begin{subfigure}[b]{0.32\textwidth}
            \centering
            \includegraphics[width=\textwidth]{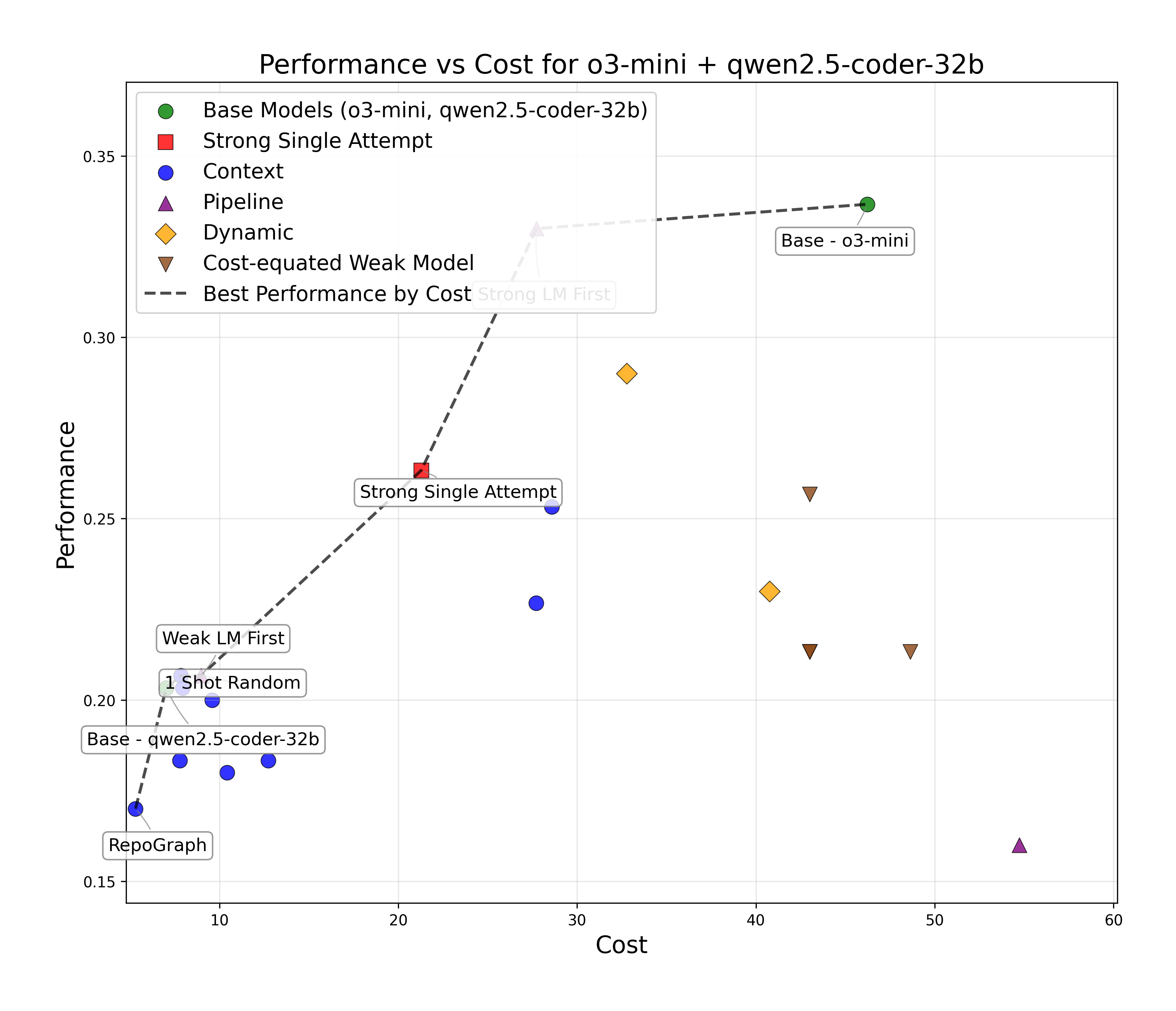}
            \caption{\centering O3-mini (0.3367) + \newline Qwen2.5-Coder-32B (0.2033)}
            \label{fig:subfig6}
        \end{subfigure} \\
    \end{tabular}
    \caption{Performance vs. cost comparison across different Strong-Weak LM pairs, denoted as (Strong LM + Weak LM). The line denotes a monotonically increasing curve, i.e. what is the best method based on performance given a particular cost budget.}
    \label{fig:perf_cost_comparison}
\end{figure*}

\begin{table*}[!h]
\centering
\resizebox{\textwidth}{!}{%
\begin{tabular}{@{}lrcrrr@{}}
\toprule
\textbf{Experiment}            & \multicolumn{1}{l}{\textbf{Resolution Rate}} & \textbf{Avg. \#iterations}                     & \multicolumn{1}{l}{\textbf{Valid Patch Rate}} & \multicolumn{1}{l}{\textbf{Total Generation Cost}} & \multicolumn{1}{l}{\textbf{Efficiency}} \\ \midrule
Base - o4-mini                 & 0.4533                                       & \textbf{1.32s + 0.00w}                         & 0.993                                         & 33.328                                             & 0.0136                                  \\
Base - gpt-4o-mini             & 0.1533                                       & 0.00s + 2.09w                                  & 0.960                                         & 4.529                                              & 0.0338                                  \\ \midrule
Self-Consistency - Direct      & {\color[HTML]{CC0000} \textbf{0.1433}}       & {\color[HTML]{CC0000} \textbf{0.00s + 15.58w}} & \textbf{0.997}                                & 35.307                                             & 0.0041                                  \\
Self-Consistency - Clustering  & 0.1500                                       & {\color[HTML]{CC0000} \textbf{0.00s + 15.58w}} & \textbf{0.997}                                & 35.307                                             & 0.0042                                  \\
Self-Consistency - Universal   & 0.1633                                       & {\color[HTML]{CC0000} \textbf{0.00s + 16.58w}} & \textbf{0.997}                                & 39.688                                             & 0.0041                                  \\
Best of n (cheating)           & 0.2167                                       & {\color[HTML]{CC0000} \textbf{0.00s + 15.58w}} & \textbf{0.997}                                & 35.307                                             & 0.0061                                  \\ \midrule
Plan                           & 0.2933                                       & {\color[HTML]{CC0000} \textbf{1.00s + 2.43w}}  & {\color[HTML]{CC0000} \textbf{0.897}}         & 29.679                                             & 0.0099                                  \\
Instance Level QA Pairs        & 0.2567                                       & 1.00s + 2.08w                                  & {\color[HTML]{CC0000} \textbf{0.923}}         & 25.598                                             & 0.0100                                  \\ \midrule
1 Shot Succesfull - Random     & {\color[HTML]{CC0000} \textbf{0.1367}}       & {\color[HTML]{CC0000} \textbf{0.01s + 2.12w}}  & 0.967                                         & 5.060                                              & 0.0270                                  \\
5 Shot Succesfull - Random     & {\color[HTML]{CC0000} \textbf{0.1400}}       & 0.04s + 1.85w                                  & 0.977                                         & 5.793                                              & 0.0242                                  \\
1 Shot Succesfull - Similarity & 0.1567                                       & 0.01s + 1.99w                                  & {\color[HTML]{CC0000} \textbf{0.957}}         & 4.796                                              & 0.0327                                  \\
5 Shot Succesfull - Similarity & {\color[HTML]{CC0000} \textbf{0.1367}}       & {\color[HTML]{CC0000} \textbf{0.04s + 2.15w}}  & {\color[HTML]{CC0000} \textbf{0.963}}         & 6.520                                              & 0.0210                                  \\
Repo Structure                 & {\color[HTML]{CC0000} \textbf{0.1367}}       & 1.00s + 2.08w                                  & {\color[HTML]{CC0000} \textbf{0.930}}         & 4.288                                              & 0.0319                                  \\
Repo Level QA Pairs            & 0.1467                                       & 1.00s + 1.92w                                  & 0.980                                         & 4.699                                              & 0.0312                                  \\
Repo Summary                   & {\color[HTML]{CC0000} \textbf{0.1433}}       & 1.00s + 2.02w                                  & {\color[HTML]{CC0000} \textbf{0.957}}         & 4.854                                              & 0.0295                                  \\ \midrule
Strong LM Single Attempt       & 0.3800                                       & 1.00s + 0.00w                                  & 0.793                                         & 23.616                                             & 0.0161                                  \\ \midrule
Strong LM First                & 0.4167                                       & 1.00s + 0.72w                                  & 0.960                                         & 25.341                                             & 0.0164                                  \\
Prompt Reduction               & 0.2733                                       & 4.36s + 1.00w                                  & {\color[HTML]{CC0000} \textbf{0.753}}         & \textbf{56.021}                                    & 0.0049                                  \\
Weak LM First                  & 0.1533                                       & 0.08s + 1.76w                                  & 0.977                                         & 6.061                                              & 0.0253                                  \\ \midrule
Weak Router                    & 0.2400                                       & {\color[HTML]{CC0000} \textbf{1.14s + 2.42w}}  & 0.963                                         & 17.903                                             & 0.0134                                  \\
Strong Router                  & 0.2133                                       & 2.29s + 1.36w                                  & 0.967                                         & 36.342                                             & 0.0059                                  \\ \bottomrule
\end{tabular}%
}
\caption{Results with O4-mini as strong LM and GPT-4o-mini as weak LM. {\color[HTML]{CC0000} \textbf{Red}} denotes drop. We consider a variance of 0.67\% owing to the non-determinism introduced by the growing temperature values so a drop less than or equal to this is not marked in red.}
\label{tab:results-o4-mini-gpt-4o-mini}
\end{table*}

\begin{table*}[!h]
\centering
\resizebox{\textwidth}{!}{%
\begin{tabular}{@{}lrcrrr@{}}
\toprule
\textbf{Experiment}            & \multicolumn{1}{l}{\textbf{Resolution Rate}} & \textbf{Avg. \#iterations}                     & \multicolumn{1}{l}{\textbf{Valid Patch Rate}} & \multicolumn{1}{l}{\textbf{Total Generation Cost}} & \multicolumn{1}{l}{\textbf{Efficiency}} \\ \midrule
Base - o3-mini                 & 0.3367                                       & \textbf{1.83s + 0.00w}                         & \textbf{0.977}                                & 46.224                                             & 0.0073                                  \\
Base - gpt-4o-mini             & 0.1533                                       & 0.00s + 2.09w                                  & 0.960                                         & 4.529                                              & \textbf{0.0338}                         \\ \midrule
Self-Consistency - Direct      & {\color[HTML]{CC0000} \textbf{0.1400}}       & {\color[HTML]{CC0000} \textbf{0.00s + 19.48w}} & 0.997                                         & 44.134                                             & 0.0032                                  \\
Self-Consistency - Clustering  & 0.1500                                       & {\color[HTML]{CC0000} \textbf{0.00s + 19.48w}} & 0.997                                         & 44.134                                             & 0.0034                                  \\
Self-Consistency - Universal   & 0.1600                                       & {\color[HTML]{CC0000} \textbf{0.00s + 20.48w}} & 0.997                                         & \textbf{48.524}                                    & 0.0033                                  \\
Best of n (cheating)           & 0.2200                                       & {\color[HTML]{CC0000} \textbf{0.00s + 19.48w}} & 0.997                                         & 44.134                                             & 0.0050                                  \\ \midrule
Plan                           & 0.2267                                       & 1.00s + 1.93w                                  & {\color[HTML]{CC0000} \textbf{0.957}}         & 24.182                                             & 0.0094                                  \\
Instance Level QA Pairs        & 0.1800                                       & 1.00s + 2.01w                                  & {\color[HTML]{CC0000} \textbf{0.943}}         & 25.699                                             & 0.0070                                  \\ \midrule
1 Shot Succesfull - Random     & {\color[HTML]{CC0000} \textbf{0.1333}}       & {\color[HTML]{CC0000} \textbf{0.01s + 2.19w}}  & {\color[HTML]{CC0000} \textbf{0.943}}         & 5.323                                              & 0.0250                                  \\
3 Shot Succesfull - Random     & {\color[HTML]{CC0000} \textbf{0.1433}}       & {\color[HTML]{CC0000} \textbf{0.03s + 2.21w}}  & {\color[HTML]{CC0000} \textbf{0.940}}         & 6.583                                              & 0.0218                                  \\
5 Shot Succesfull - Random     & {\color[HTML]{CC0000} \textbf{0.1267}}       & 0.05s + 2.03w                                  & 0.963                                         & 7.127                                              & 0.0178                                  \\
1 Shot Succesfull - Similarity & {\color[HTML]{CC0000} \textbf{0.1433}}       & {\color[HTML]{CC0000} \textbf{0.01s + 2.46w}}  & {\color[HTML]{CC0000} \textbf{0.923}}         & 6.081                                              & 0.0236                                  \\
3 Shot Succesfull - Similarity & {\color[HTML]{CC0000} \textbf{0.1367}}       & 0.03s + 2.01w                                  & {\color[HTML]{CC0000} \textbf{0.950}}         & 6.093                                              & 0.0224                                  \\
5 Shot Succesfull - Similarity & 0.1500                                       & {\color[HTML]{CC0000} \textbf{0.05s + 2.14w}}  & {\color[HTML]{CC0000} \textbf{0.947}}         & 7.320                                              & 0.0205                                  \\
1 Shot All - Random            & 0.1467                                       & {\color[HTML]{CC0000} \textbf{0.00s + 2.35w}}  & {\color[HTML]{CC0000} \textbf{0.927}}         & 5.716                                              & 0.0257                                  \\
3 Shot All - Random            & {\color[HTML]{CC0000} \textbf{0.1333}}       & {\color[HTML]{CC0000} \textbf{0.01s + 2.19w}}  & {\color[HTML]{CC0000} \textbf{0.937}}         & 5.366                                              & 0.0248                                  \\
5 Shot All - Random            & 0.1500                                       & {\color[HTML]{CC0000} \textbf{0.02s + 2.10w}}  & {\color[HTML]{CC0000} \textbf{0.950}}         & 5.712                                              & 0.0263                                  \\
1 Shot All - Similarity        & {\color[HTML]{CC0000} \textbf{0.1200}}       & {\color[HTML]{CC0000} \textbf{0.00s + 2.44w}}  & {\color[HTML]{CC0000} \textbf{0.923}}         & 5.627                                              & 0.0213                                  \\
3 Shot All - Similarity        & {\color[HTML]{CC0000} \textbf{0.1367}}       & {\color[HTML]{CC0000} \textbf{0.01s + 2.37w}}  & {\color[HTML]{CC0000} \textbf{0.930}}         & 5.959                                              & 0.0229                                  \\
5 Shot All - Similarity        & 0.1500                                       & {\color[HTML]{CC0000} \textbf{0.02s + 2.19w}}  & 0.960                                         & 5.818                                              & 0.0258                                  \\
Repo Structure                 & {\color[HTML]{CC0000} \textbf{0.1367}}       & 1.00s + 2.08w                                  & {\color[HTML]{CC0000} \textbf{0.930}}         & 4.288                                              & 0.0319                                  \\
Repo Level QA Pairs            & {\color[HTML]{CC0000} \textbf{0.1267}}       & 1.00s + 1.86w                                  & 0.977                                         & 4.686                                              & 0.0270                                  \\
Repo Summary                   & 0.1633                                       & 1.00s + 2.07w                                  & 0.960                                         & 5.000                                              & 0.0327                                  \\ \midrule
Strong LM Single Attempt       & 0.2633                                       & 1.00s + 0.00w                                  & {\color[HTML]{CC0000} \textbf{0.690}}         & 21.280                                             & 0.0124                                  \\ \midrule
Strong LM First                & 0.3000                                       & 1.00s + 1.17w                                  & {\color[HTML]{CC0000} \textbf{0.930}}         & 24.922                                             & 0.0120                                  \\
Prompt Reduction               & 0.2033                                       & 4.73s + 1.00w                                  & {\color[HTML]{CC0000} \textbf{0.653}}         & 42.537                                             & 0.0048                                  \\
Weak LM First                  & 0.1633                                       & 0.08s + 1.61w                                  & 0.990                                         & 5.687                                              & 0.0287                                  \\ \midrule
Weak Router                    & 0.2267                                       & 0.48s + 2.30w                                  & 0.977                                         & 16.764                                             & 0.0135                                  \\
Strong Router                  & 0.1867                                       & 1.71s + 1.22w                                  & 0.977                                         & 36.583                                             & 0.0051                                  \\ \bottomrule
\end{tabular}%
}
\caption{Results with O3-mini as strong LM and GPT-4o-mini as weak LM. {\color[HTML]{CC0000} \textbf{Red}} denotes drop. We consider a variance of 0.67\% owing to the non-determinism introduced by the growing temperature values so a drop less than or equal to this is not marked in red.}
\label{tab:results-o3-mini-gpt-4o-mini}
\end{table*}

\begin{table*}[]
\centering
\resizebox{\textwidth}{!}{%
\begin{tabular}{@{}lrcrrr@{}}
\toprule
\textbf{Experiment}            & \multicolumn{1}{l}{\textbf{Resolution Rate}} & \textbf{Avg. \#iterations}                     & \multicolumn{1}{l}{\textbf{Valid Patch Rate}} & \multicolumn{1}{l}{\textbf{Total Generation Cost}} & \multicolumn{1}{l}{\textbf{Efficiency}} \\ \midrule
Base - o3-mini                 & 0.3367                                       & \textbf{1.83s + 0.00w}                         & \textbf{0.977}                                & 46.224                                             & 0.0073                                  \\
Base - qwen2.5-coder-7b        & 0.0467                                       & 0.00s + 6.48w                                  & 0.600                                         & 3.839                                              & 0.0122                                  \\ \midrule
Self-Consistency - Direct      & 0.0733                                       & {\color[HTML]{CC0000} \textbf{0.00s + 78.38w}} & 0.877                                         & 46.090                                             & 0.0016                                  \\
Self-Consistency - Clustering  & 0.0700                                       & {\color[HTML]{CC0000} \textbf{0.00s + 78.38w}} & 0.877                                         & 46.090                                             & 0.0015                                  \\
Self-Consistency - Universal   & 0.0733                                       & {\color[HTML]{CC0000} \textbf{0.00s + 79.38w}} & 0.877                                         & \textbf{47.095}                                    & 0.0016                                  \\
Best of n (cheating)           & 0.1033                                       & {\color[HTML]{CC0000} \textbf{0.00s + 78.38w}} & 0.877                                         & 46.090                                             & 0.0022                                  \\ \midrule
Plan                           & 0.0833                                       & {\color[HTML]{CC0000} \textbf{1.00s + 6.79w}}  & {\color[HTML]{CC0000} \textbf{0.517}}         & 24.092                                             & 0.0035                                  \\
Instance Level QA Pairs        & 0.0767                                       & 1.00s + 6.95w                                  & {\color[HTML]{CC0000} \textbf{0.563}}         & 25.203                                             & 0.0030                                  \\ \midrule
1 Shot Succesfull - Random     & 0.0433                                       & {\color[HTML]{CC0000} \textbf{0.01s + 6.49w}}  & 0.617                                         & 4.281                                              & 0.0101                                  \\
5 Shot Succesfull - Random     & 0.0433                                       & {\color[HTML]{CC0000} \textbf{0.05s + 6.61w}}  & {\color[HTML]{CC0000} \textbf{0.590}}         & 6.501                                              & 0.0067                                  \\
1 Shot Succesfull - Similarity & {\color[HTML]{CC0000} \textbf{0.0367}}       & {\color[HTML]{CC0000} \textbf{0.01s + 6.63w}}  & 0.603                                         & 4.282                                              & 0.0086                                  \\
5 Shot Succesfull - Similarity & 0.0433                                       & {\color[HTML]{CC0000} \textbf{0.05s + 6.86w}}  & {\color[HTML]{CC0000} \textbf{0.570}}         & 6.665                                              & 0.0065                                  \\
Repo Structure                 & 0.0500                                       & {\color[HTML]{CC0000} \textbf{1.00s + 7.32w}}  & {\color[HTML]{CC0000} \textbf{0.487}}         & 4.388                                              & 0.0114                                  \\
Repo Level QA Pairs            & 0.0433                                       & {\color[HTML]{CC0000} \textbf{1.00s + 6.75w}}  & {\color[HTML]{CC0000} \textbf{0.577}}         & 4.676                                              & 0.0093                                  \\
Repo Summary                   & 0.0567                                       & {\color[HTML]{CC0000} \textbf{1.00s + 6.70w}}  & {\color[HTML]{CC0000} \textbf{0.587}}         & 9.139                                              & 0.0062                                  \\ \midrule
Strong LM Single Attempt       & 0.2633                                       & 1.00s + 0.00w                                  & {\color[HTML]{CC0000} \textbf{0.690}}         & 21.280                                             & \textbf{0.0124}                         \\ \midrule
Strong LM First                & 0.2633                                       & 1.00s + 2.66w                                  & 0.773                                         & 22.261                                             & 0.0118                                  \\
Prompt Reduction               & 0.1867                                       & 4.56s + 1.00w                                  & 0.663                                         & 43.590                                             & 0.0043                                  \\
Weak LM First                  & 0.1933                                       & 0.62s + 3.82w                                  & 0.810                                         & 16.147                                             & 0.0120                                  \\ \midrule
Weak Router                    & 0.0733                                       & {\color[HTML]{CC0000} \textbf{0.07s + 7.42w}}  & 0.607                                         & 5.432                                              & 0.0135                                  \\
Strong Router                  & 0.1033                                       & 1.77s + 3.72w                                  & 0.820                                         & 41.081                                             & 0.0025                                  \\ \bottomrule
\end{tabular}%
}
\caption{Results with O3-mini as strong LM and Qwen2.5-Coder-7B as weak LM. {\color[HTML]{CC0000} \textbf{Red}} denotes drop. We consider a variance of 0.67\% owing to the non-determinism introduced by the growing temperature values so a drop less than or equal to this is not marked in red.}
\label{tab:results-o3-mini-qwen25coder7b}
\end{table*}

\begin{table*}[]
\centering
\resizebox{\textwidth}{!}{%
\begin{tabular}{@{}lrcrrr@{}}
\toprule
\textbf{Experiment}            & \multicolumn{1}{l}{\textbf{Resolution Rate}} & \textbf{Avg. \#iterations}                     & \multicolumn{1}{l}{\textbf{Valid Patch Rate}} & \multicolumn{1}{l}{\textbf{Total Generation Cost}} & \multicolumn{1}{l}{\textbf{Efficiency}} \\ \midrule
Base - o3-mini                 & 0.3367                                       & \textbf{1.83s + 0.00w}                         & 0.977                                         & 46.224                                             & 0.0073                                  \\
Base - qwen2.5-coder-14b       & 0.1433                                       & 0.00s + 2.91w                                  & 0.937                                         & 4.257                                              & \textbf{0.0337}                         \\ \midrule
Self-Consistency - Direct      & 0.1500                                       & {\color[HTML]{CC0000} \textbf{0.00s + 32.59w}} & \textbf{0.987}                                & 48.249                                             & 0.0031                                  \\
Self-Consistency - Clustering  & 0.1567                                       & {\color[HTML]{CC0000} \textbf{0.00s + 32.59w}} & \textbf{0.987}                                & 48.249                                             & 0.0032                                  \\
Self-Consistency - Universal   & 0.1500                                       & {\color[HTML]{CC0000} \textbf{0.00s + 33.59w}} & \textbf{0.987}                                & \textbf{51.069}                                    & 0.0029                                  \\
Best of n (cheating)           & 0.2333                                       & {\color[HTML]{CC0000} \textbf{0.00s + 32.59w}} & \textbf{0.987}                                & 48.249                                             & 0.0048                                  \\ \midrule
Plan                           & 0.2233                                       & {\color[HTML]{CC0000} \textbf{1.00s + 3.21w}}  & {\color[HTML]{CC0000} \textbf{0.893}}         & 25.158                                             & 0.0089                                  \\
Instance Level QA Pairs        & 0.1900                                       & 1.00s + 2.82w                                  & {\color[HTML]{CC0000} \textbf{0.927}}         & 25.599                                             & 0.0074                                  \\ \midrule
1 Shot Succesfull - Random     & {\color[HTML]{CC0000} \textbf{0.1233}}       & {\color[HTML]{CC0000} \textbf{0.01s + 3.78w}}  & {\color[HTML]{CC0000} \textbf{0.863}}         & 6.140                                              & 0.0201                                  \\
5 Shot Succesfull - Random     & {\color[HTML]{CC0000} \textbf{0.1033}}       & {\color[HTML]{CC0000} \textbf{0.05s + 4.15w}}  & {\color[HTML]{CC0000} \textbf{0.830}}         & 8.641                                              & 0.0120                                  \\
1 Shot Succesfull - Similarity & {\color[HTML]{CC0000} \textbf{0.1200}}       & {\color[HTML]{CC0000} \textbf{0.01s + 3.92w}}  & {\color[HTML]{CC0000} \textbf{0.880}}         & 6.427                                              & 0.0187                                  \\
5 Shot Succesfull - Similarity & {\color[HTML]{CC0000} \textbf{0.1233}}       & {\color[HTML]{CC0000} \textbf{0.05s + 3.97w}}  & {\color[HTML]{CC0000} \textbf{0.860}}         & 8.713                                              & 0.0142                                  \\
Repo Structure                 & 0.1400                                       & {\color[HTML]{CC0000} \textbf{1.00s + 3.04w}}  & {\color[HTML]{CC0000} \textbf{0.880}}         & 4.393                                              & 0.0319                                  \\
Repo Level QA Pairs            & 0.1400                                       & {\color[HTML]{CC0000} \textbf{1.00s + 2.99w}}  & {\color[HTML]{CC0000} \textbf{0.907}}         & 5.036                                              & 0.0278                                  \\
Repo Summary                   & 0.1467                                       & {\color[HTML]{CC0000} \textbf{1.00s + 3.12w}}  & {\color[HTML]{CC0000} \textbf{0.930}}         & 9.609                                              & 0.0153                                  \\ \midrule
Strong LM Single Attempt       & 0.2633                                       & 1.00s + 0.00w                                  & {\color[HTML]{CC0000} \textbf{0.690}}         & 21.280                                             & 0.0124                                  \\ \midrule
Strong LM First                & 0.2967                                       & 1.00s + 2.20w                                  & {\color[HTML]{CC0000} \textbf{0.830}}         & 25.099                                             & 0.0118                                  \\
Prompt Reduction               & 0.2300                                       & 4.36s + 1.00w                                  & {\color[HTML]{CC0000} \textbf{0.707}}         & 40.568                                             & 0.0057                                  \\
Weak LM First                  & 0.1667                                       & 0.20s + 2.72w                                  & {\color[HTML]{CC0000} \textbf{0.913}}         & 7.847                                              & 0.0212                                  \\ \midrule
Weak Router                    & 0.2400                                       & 1.34s + 2.45w                                  & 0.947                                         & 27.507                                             & 0.0087                                  \\
Strong Router                  & 0.1633                                       & 2.32s + 1.68w                                  & 0.937                                         & 39.201                                             & 0.0042                                  \\ \bottomrule
\end{tabular}%
}
\caption{Results with O3-mini as strong LM and Qwen2.5-Coder-14B as weak LM. {\color[HTML]{CC0000} \textbf{Red}} denotes drop. We consider a variance of 0.67\% owing to the non-determinism introduced by the growing temperature values so a drop less than or equal to this is not marked in red.}
\label{tab:results-o3-mini-qwen25coder14b}
\end{table*}

\begin{table*}[]
\centering
\resizebox{\textwidth}{!}{%
\begin{tabular}{@{}lrcrrr@{}}
\toprule
\textbf{Experiment}            & \multicolumn{1}{l}{\textbf{Resolution Rate}} & \textbf{Avg. \#iterations}                     & \multicolumn{1}{l}{\textbf{Valid Patch Rate}} & \multicolumn{1}{l}{\textbf{Total Generation Cost}} & \multicolumn{1}{l}{\textbf{Efficiency}} \\ \midrule
Base - o3-mini                 & 0.3367                                       & \textbf{1.83s + 0.00w}                         & 0.977                                         & 46.224                                             & 0.0073                                  \\
Base - qwen2.5-coder-32b       & 0.2033                                       & 0.00s + 2.35w                                  & 0.967                                         & 7.021                                              & 0.0290                                  \\ \midrule
Self-Consistency - Direct      & 0.2133                                       & {\color[HTML]{CC0000} \textbf{0.00s + 12.98w}} & \textbf{0.987}                                & 43.011                                             & 0.0050                                  \\
Self-Consistency - Clustering  & 0.2133                                       & {\color[HTML]{CC0000} \textbf{0.00s + 12.98w}} & \textbf{0.987}                                & 43.011                                             & 0.0050                                  \\
Self-Consistency - Universal   & 0.2133                                       & {\color[HTML]{CC0000} \textbf{0.00s + 13.98w}} & \textbf{0.987}                                & \textbf{48.634}                                    & 0.0044                                  \\
Best of n                      & 0.2567                                       & {\color[HTML]{CC0000} \textbf{0.00s + 12.98w}} & \textbf{0.987}                                & 43.011                                             & 0.0060                                  \\ \midrule
Plan                           & 0.2533                                       & {\color[HTML]{CC0000} \textbf{1.00s + 2.64w}}  & {\color[HTML]{CC0000} \textbf{0.913}}         & 28.591                                             & 0.0089                                  \\
Instance Level QA Pairs        & 0.2267                                       & 1.00s + 2.22w                                  & {\color[HTML]{CC0000} \textbf{0.957}}         & 27.700                                             & 0.0082                                  \\ \midrule
1 Shot Succesfull - Random     & 0.2067                                       & {\color[HTML]{CC0000} \textbf{0.01s + 2.38w}}  & {\color[HTML]{CC0000} \textbf{0.957}}         & 7.843                                              & 0.0264                                  \\
5 Shot Succesfull - Random     & {\color[HTML]{CC0000} \textbf{0.1800}}       & {\color[HTML]{CC0000} \textbf{0.05s + 2.35w}}  & {\color[HTML]{CC0000} \textbf{0.940}}         & 10.426                                             & 0.0173                                  \\
1 Shot Succesfull - Similarity & 0.2033                                       & {\color[HTML]{CC0000} \textbf{0.01s + 2.46w}}  & {\color[HTML]{CC0000} \textbf{0.947}}         & 7.933                                              & 0.0256                                  \\
5 Shot Succesfull - Similarity & 0.2000                                       & 0.05s + 2.34w                                  & 0.967                                         & 9.583                                              & 0.0209                                  \\
Repo Structure                 & {\color[HTML]{CC0000} \textbf{0.1700}}       & {\color[HTML]{CC0000} \textbf{1.00s + 2.63w}}  & {\color[HTML]{CC0000} \textbf{0.887}}         & 5.302                                              & \textbf{0.0321}                         \\
Repo Level QA Pairs            & {\color[HTML]{CC0000} \textbf{0.1833}}       & 1.00s + 2.33w                                  & {\color[HTML]{CC0000} \textbf{0.950}}         & 7.785                                              & 0.0235                                  \\
Repo Summary                   & {\color[HTML]{CC0000} \textbf{0.1833}}       & {\color[HTML]{CC0000} \textbf{1.00s + 2.47w}}  & {\color[HTML]{CC0000} \textbf{0.933}}         & 12.730                                             & 0.0144                                  \\ \midrule
Strong LM Single Attempt       & 0.2633                                       & 1.00s + 0.00w                                  & {\color[HTML]{CC0000} \textbf{0.690}}         & 21.280                                             & 0.0124                                  \\ \midrule
Strong LM First                & 0.3300                                       & 1.00s + 1.80w                                  & {\color[HTML]{CC0000} \textbf{0.870}}         & 27.737                                             & 0.0119                                  \\
Prompt Reduction               & {\color[HTML]{CC0000} \textbf{0.1600}}       & 5.63s + 1.00w                                  & {\color[HTML]{CC0000} \textbf{0.557}}         & 54.735                                             & 0.0029                                  \\
Weak LM First                  & 0.2067                                       & 0.12s + 2.31w                                  & {\color[HTML]{CC0000} \textbf{0.930}}         & 8.9637                                             & 0.0231                                  \\ \midrule
Weak Router                    & 0.2900                                       & 1.42s + 1.83w                                  & 0.970                                         & 32.772                                             & 0.0088                                  \\
Strong Router                  & 0.2300                                       & 2.31s + 1.50w                                  & 0.973                                         & 40.763                                             & 0.0056                                  \\ \bottomrule
\end{tabular}%
}
\caption{Results with O3-mini as strong LM and Qwen2.5-Coder-32B as weak LM. {\color[HTML]{CC0000} \textbf{Red}} denotes drop. We consider a variance of 0.67\% owing to the non-determinism introduced by the growing temperature values so a drop less than or equal to this is not marked in red.}
\label{tab:results-o3-mini-qwen25coder32b}
\end{table*}

\begin{table*}[]
\centering
\resizebox{\textwidth}{!}{%
\begin{tabular}{@{}lrcrrr@{}}
\toprule
\textbf{Experiment}            & \multicolumn{1}{l}{\textbf{Resolution Rate}} & \textbf{Avg. \#iterations}                    & \multicolumn{1}{l}{\textbf{Valid Patch Rate}} & \multicolumn{1}{l}{\textbf{Total Generation Cost}} & \multicolumn{1}{l}{\textbf{Efficiency}} \\ \midrule
Base - gpt-4o-mini             & \textbf{0.1533}                              & \textbf{2.09s + 0.00w}                        & \textbf{0.960}                                & 4.529                                              & 0.0338                                  \\
Base - qwen2.5-coder-7b        & 0.0467                                       & 0.00s + 6.48w                                 & 0.600                                         & 3.839                                              & 0.0122                                  \\ \midrule
Self-Consistency - Direct      & 0.0467                                       & 0.00s + 6.48w                                 & 0.600                                         & 3.839                                              & 0.0122                                  \\
Self-Consistency - Clustering  & 0.0467                                       & 0.00s + 6.48w                                 & 0.600                                         & 3.839                                              & 0.0122                                  \\
Self-Consistency - Universal   & 0.0467                                       & 0.00s + 6.48w                                 & 0.600                                         & 3.839                                              & 0.0122                                  \\
Best of n (cheating)           & 0.0467                                       & 0.00s + 6.48w                                 & 0.600                                         & 3.839                                              & 0.0122                                  \\ \midrule
Plan                           & 0.0467                                       & {\color[HTML]{CC0000} \textbf{1.00s + 7.14w}} & {\color[HTML]{CC0000} \textbf{0.483}}         & 6.262                                              & 0.0075                                  \\
Instance Level QA Pairs        & 0.0600                                       & {\color[HTML]{CC0000} \textbf{1.00s + 6.99w}} & {\color[HTML]{CC0000} \textbf{0.533}}         & \textbf{6.402}                                     & 0.0094                                  \\ \midrule
1 Shot Succesfull - Random     & 0.0500                                       & {\color[HTML]{CC0000} \textbf{0.02s + 6.25w}} & 0.633                                         & 3.805                                              & 0.0131                                  \\
5 Shot Succesfull - Random     & 0.0533                                       & {\color[HTML]{CC0000} \textbf{0.10s + 6.90w}} & {\color[HTML]{CC0000} \textbf{0.547}}         & 4.854                                              & 0.0110                                  \\
1 Shot Succesfull - Similarity & 0.0433                                       & {\color[HTML]{CC0000} \textbf{0.02s + 6.58w}} & {\color[HTML]{CC0000} \textbf{0.590}}         & 4.284                                              & 0.0101                                  \\
5 Shot Succesfull - Similarity & 0.0400                                       & {\color[HTML]{CC0000} \textbf{0.10s + 6.79w}} & {\color[HTML]{CC0000} \textbf{0.553}}         & 5.733                                              & 0.0070                                  \\
Repo Structure                 & 0.0467                                       & {\color[HTML]{CC0000} \textbf{1.00s + 7.42w}} & {\color[HTML]{CC0000} \textbf{0.483}}         & 1.562                                              & 0.0299                                  \\
Repo Level QA Pairs            & {\color[HTML]{CC0000} \textbf{0.0367}}       & {\color[HTML]{CC0000} \textbf{1.00s + 6.73w}} & {\color[HTML]{CC0000} \textbf{0.563}}         & 4.105                                              & 0.0089                                  \\
Repo Summary                   & 0.0433                                       & {\color[HTML]{CC0000} \textbf{1.00s + 6.63w}} & {\color[HTML]{CC0000} \textbf{0.570}}         & 3.873                                              & 0.0112                                  \\ \midrule
Strong LM Single Attempt       & 0.1233                                       & 1.00s + 0.00w                                 & 0.693                                         & 2.046                                              & \textbf{0.0603}                         \\ \midrule
Strong LM First                & 0.1233                                       & 1.00s + 3.28w                                 & 0.770                                         & 3.579                                              & 0.0344                                  \\
Prompt Reduction               & 0.0567                                       & 4.83s + 1.00w                                 & 0.647                                         & 1.246                                              & 0.0455                                  \\
Weak LM First                  & 0.1133                                       & 0.18s + 2.86w                                 & 0.823                                         & 3.581                                              & 0.0316                                  \\ \midrule
Weak Router                    & 0.0533                                       & {\color[HTML]{CC0000} \textbf{0.62s + 7.26w}} & 0.623                                         & 4.326                                              & 0.0123                                  \\
Strong Router                  & 0.0700                                       & 1.72s + 5.14w                                 & 0.723                                         & 5.941                                              & 0.0118                                  \\ \bottomrule
\end{tabular}%
}
\caption{Results with GPT-4o-mini as strong LM and Qwen2.5-Coder-7B as weak LM. {\color[HTML]{CC0000} \textbf{Red}} denotes drop. We consider a variance of 0.67\% owing to the non-determinism introduced by the growing temperature values so a drop less than or equal to this is not marked in red.}
\label{tab:results-gpt-4o-mini-qwen25coder7b}
\end{table*}


\begin{figure*}[!h]
    \centering
    \begin{subfigure}[b]{0.3\textwidth}
        \centering
        \includegraphics[width=\textwidth]{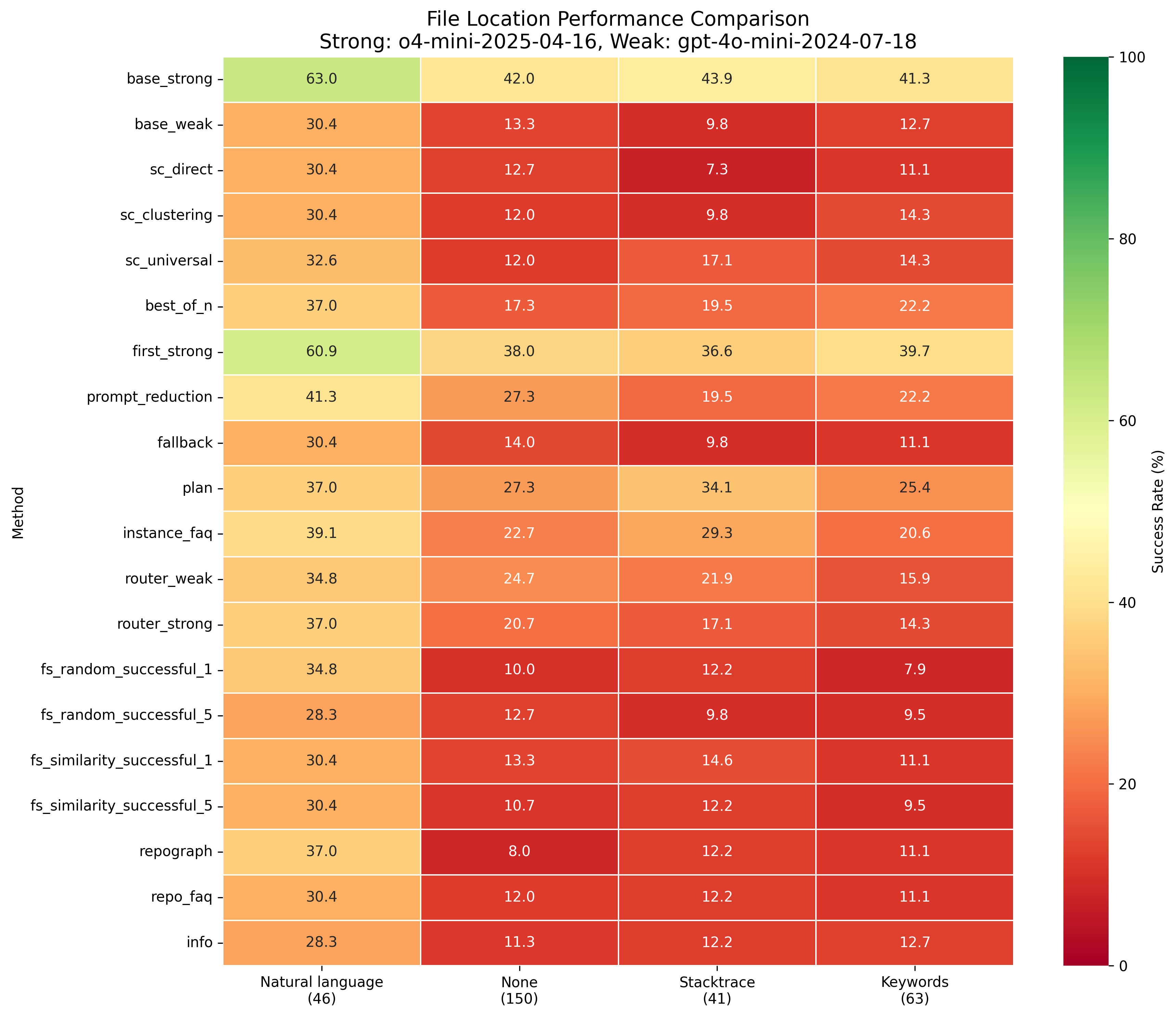}
        \caption{File Location}
        \label{fig:fig1}
    \end{subfigure}
    \hfill
    \begin{subfigure}[b]{0.3\textwidth}
        \centering
        \includegraphics[width=\textwidth]{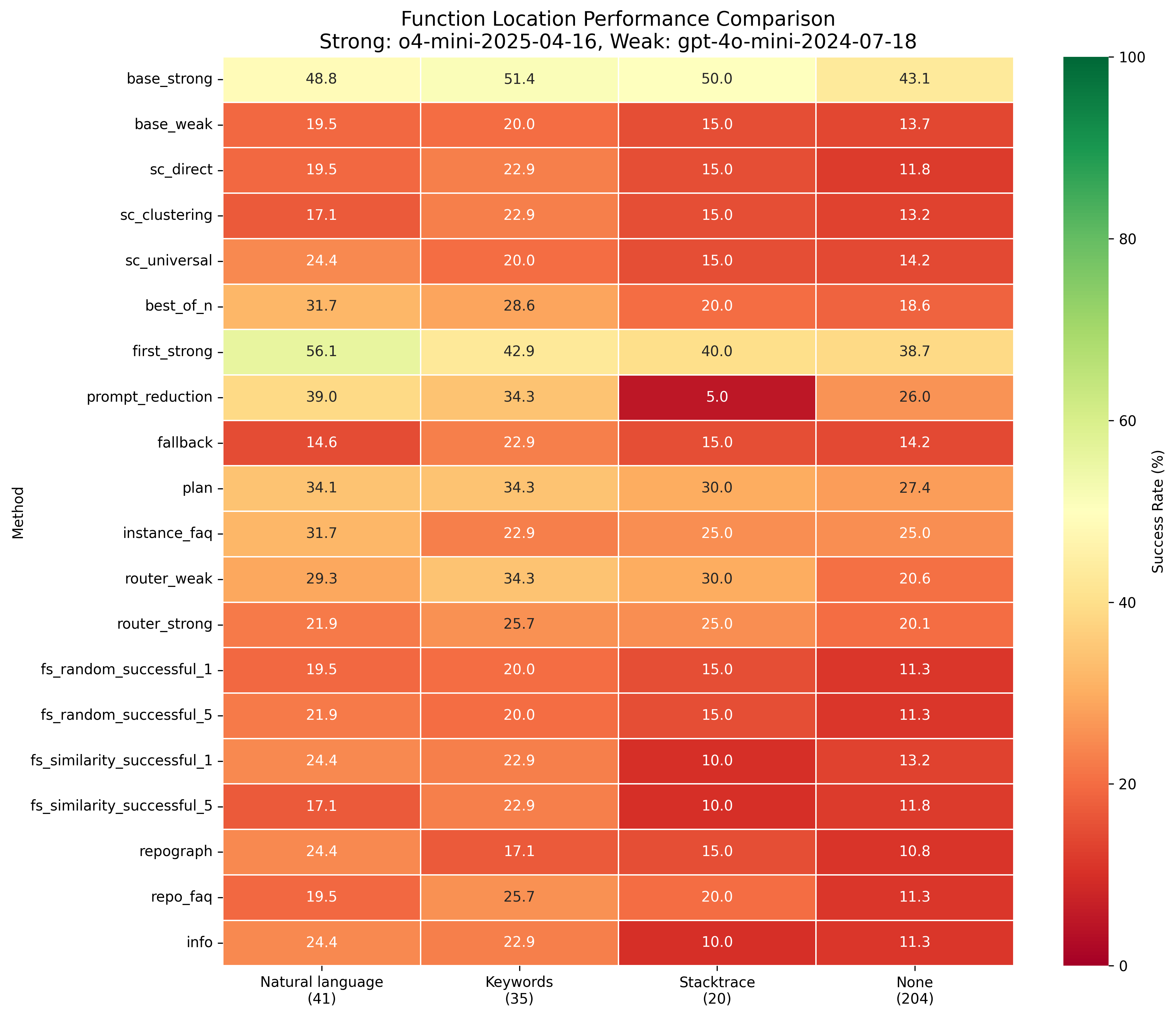}
        \caption{Function Location}
        \label{fig:fig2}
    \end{subfigure}
    \hfill
    \begin{subfigure}[b]{0.3\textwidth}
        \centering
        \includegraphics[width=\textwidth]{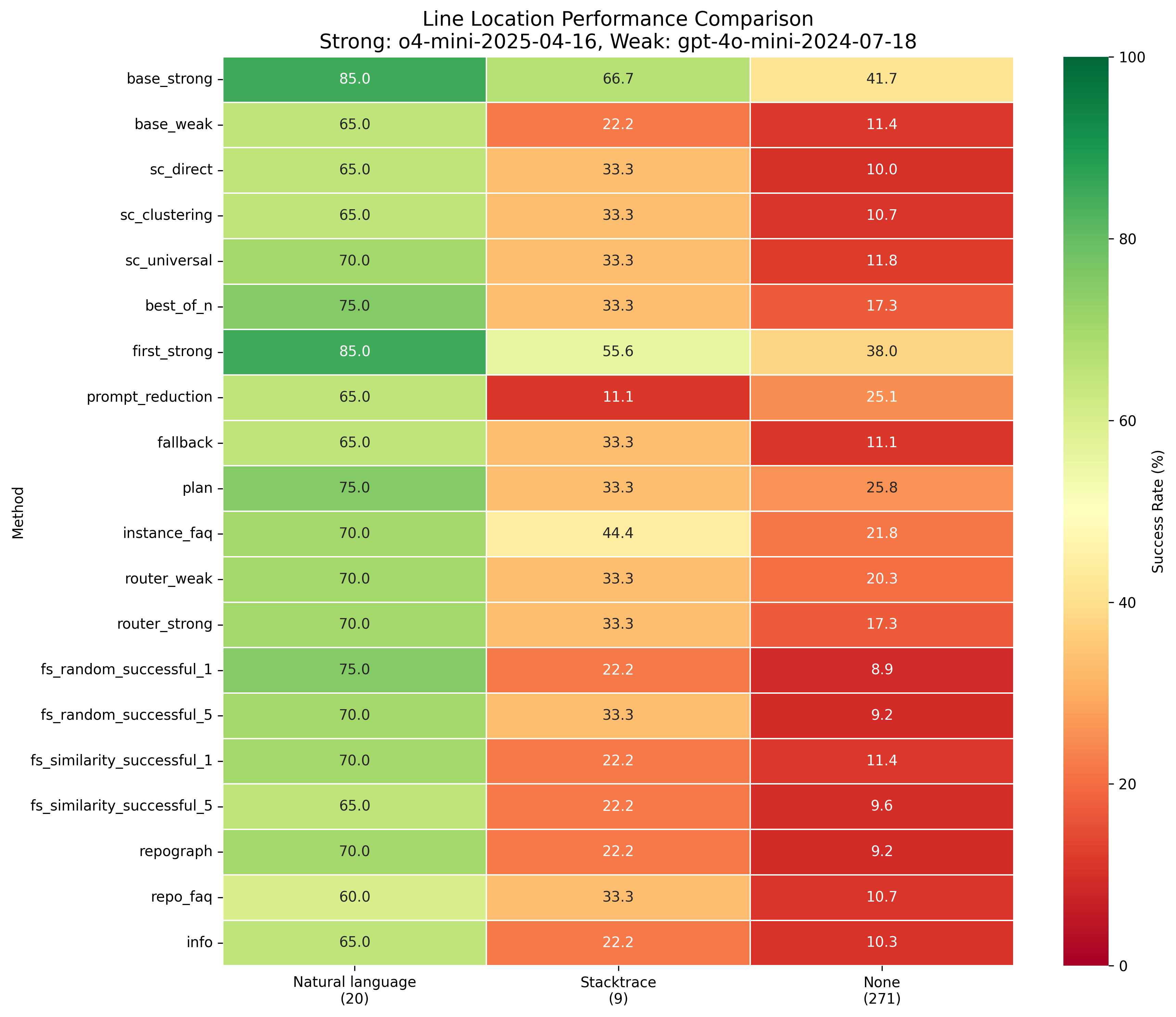}
        \caption{Line Location}
        \label{fig:fig3}
    \end{subfigure}

    \vspace{1em}
    
    \begin{minipage}{\textwidth}
        \centering
        \begin{subfigure}[b]{0.3\textwidth}
            \centering
            \includegraphics[width=\textwidth]{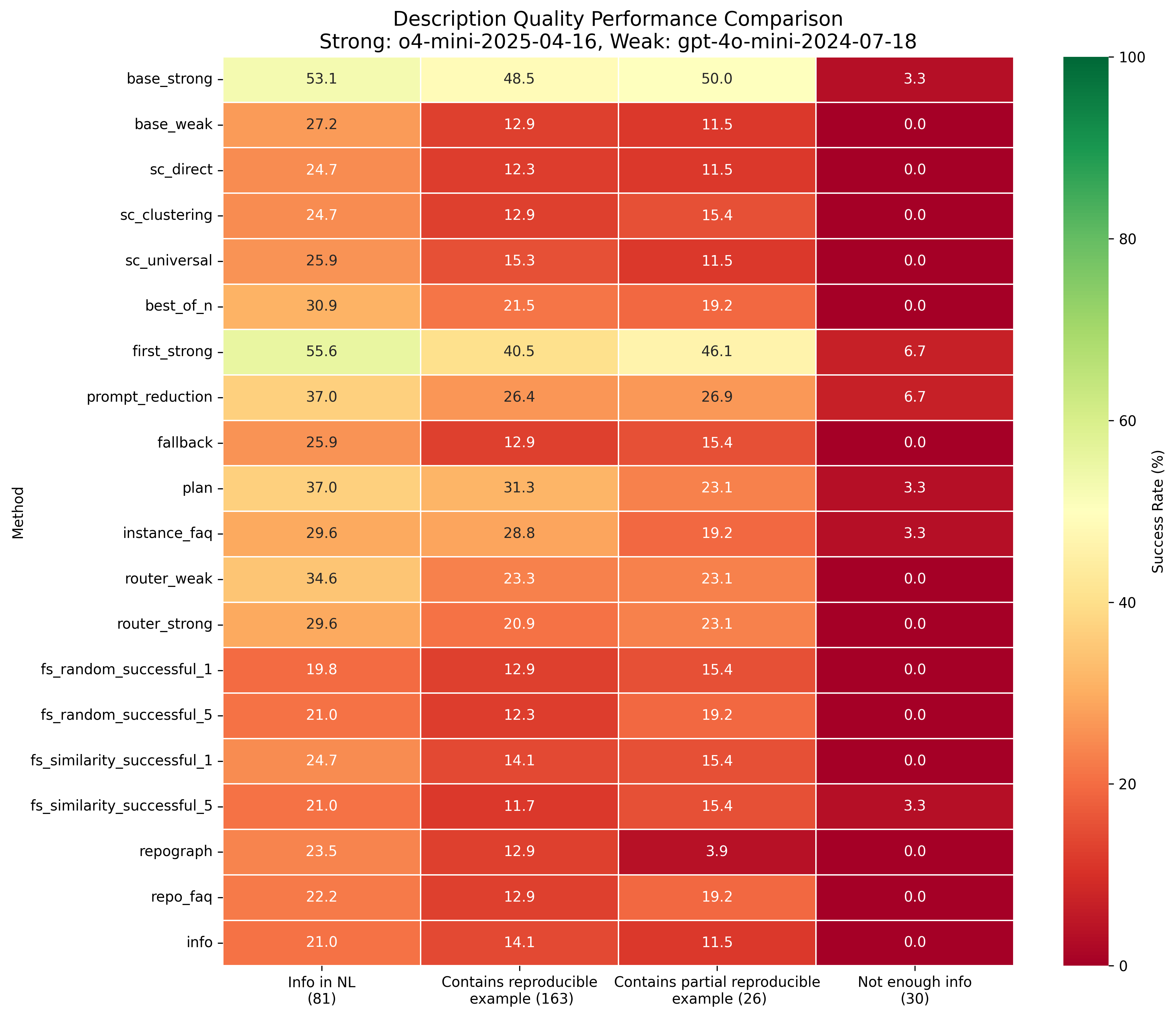}
            \caption{Description Quality}
            \label{fig:fig4}
        \end{subfigure}
        \quad
        \begin{subfigure}[b]{0.3\textwidth}
            \centering
            \includegraphics[width=\textwidth]{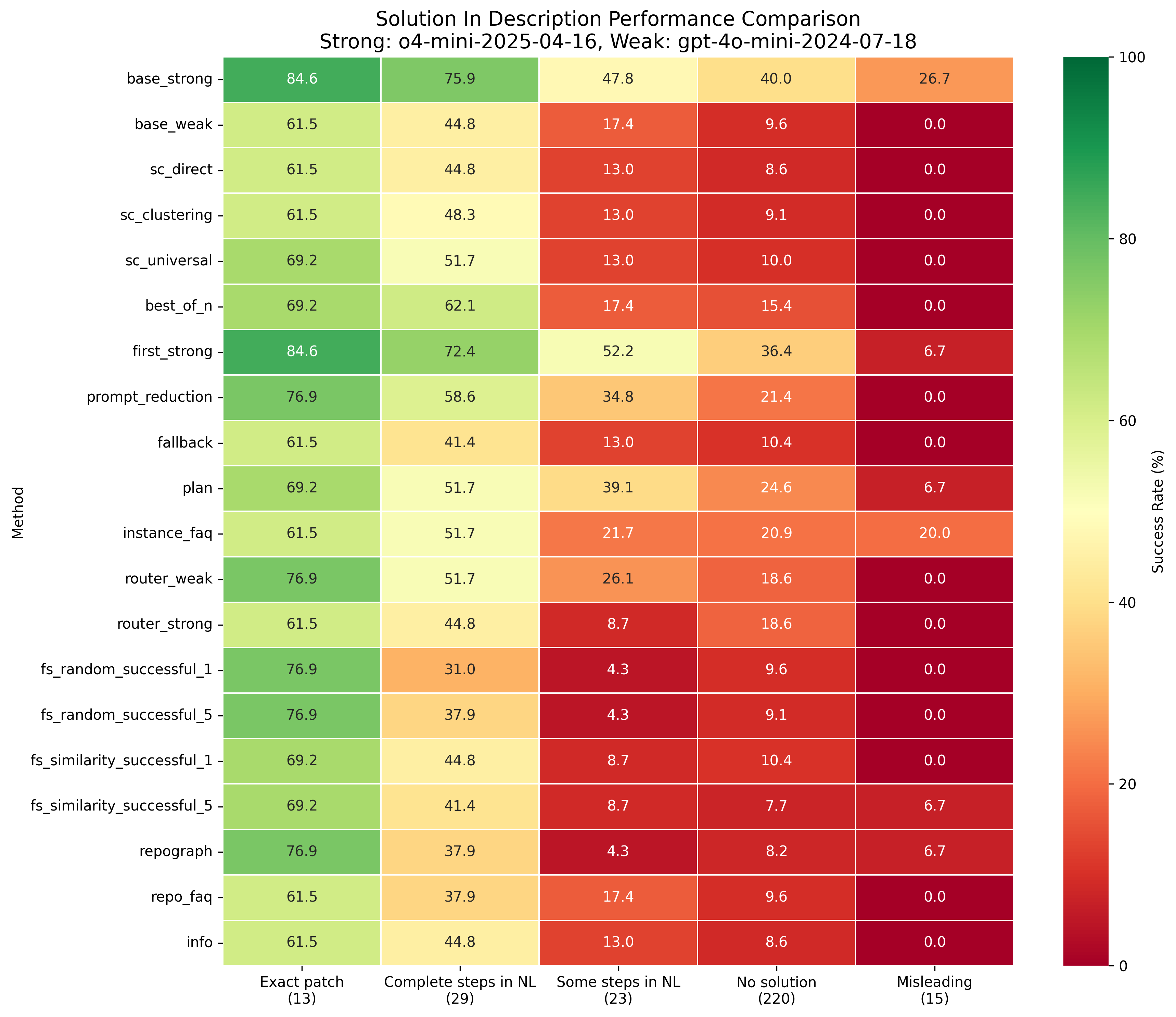}
            \caption{Solution in Description}
            \label{fig:fig5}
        \end{subfigure}
    \end{minipage}
    
    \caption{Heatmaps of issue category wise performance for O4-mini + GPT-4o-mini model pair. }
    \label{fig:issue_category_wise}
\end{figure*}


\begin{table*}[!h]
\centering
\resizebox{\textwidth}{!}{%
\begin{tabular}{@{}lrrrrrrrrr@{}}
\toprule
\textbf{Method}                & \multicolumn{1}{l}{\textbf{File Precision}} & \multicolumn{1}{l}{\textbf{File Recall}} & \multicolumn{1}{l}{\textbf{File F1}} & \multicolumn{1}{l}{\textbf{Module Precision}} & \multicolumn{1}{l}{\textbf{Module Recall}} & \multicolumn{1}{l}{\textbf{Module F1}} & \multicolumn{1}{l}{\textbf{Line Precision}} & \multicolumn{1}{l}{\textbf{Line Recall}} & \multicolumn{1}{l}{\textbf{Line F1}} \\ \midrule
Base Strong                    & 84.23                                       & 84.23                                    & 84.23                                & 84.23                                         & 84.23                                      & 84.23                                  & 55.13                                       & 47.33                                    & 48.43                                \\
Base Weak                      & 78.65                                       & 78.82                                    & 78.7                                 & 78.65                                         & 78.82                                      & 78.7                                   & 35.92                                       & 28.49                                    & 29.97                                \\
Self-Consistency - Direct      & \cellcolor[HTML]{F4CCCC}76.92               & \cellcolor[HTML]{F4CCCC}76.92            & \cellcolor[HTML]{F4CCCC}76.92        & \cellcolor[HTML]{F4CCCC}76.92                 & \cellcolor[HTML]{F4CCCC}76.92              & \cellcolor[HTML]{F4CCCC}76.92          & \cellcolor[HTML]{F4CCCC}35.56               & \cellcolor[HTML]{F4CCCC}27.74            & \cellcolor[HTML]{F4CCCC}29.21        \\
\bottomrule
\end{tabular}%
}
\caption{Localization performance for O4-mini as strong LM and GPT-4o-mini as weak LM. Cells highlighted in red depict lower score than the base weak LM setting.}
\label{tab:localization_o4-mini_gpt-4o-mini}
\end{table*}
\begin{table*}[!h]
\centering
\resizebox{\textwidth}{!}{%
\begin{tabular}{@{}lrrrrrrrrr@{}}
\toprule
\textbf{Method}                & \multicolumn{1}{l}{\textbf{File Precision}} & \multicolumn{1}{l}{\textbf{File Recall}} & \multicolumn{1}{l}{\textbf{File F1}} & \multicolumn{1}{l}{\textbf{Module Precision}} & \multicolumn{1}{l}{\textbf{Module Recall}} & \multicolumn{1}{l}{\textbf{Module F1}} & \multicolumn{1}{l}{\textbf{Line Precision}} & \multicolumn{1}{l}{\textbf{Line Recall}} & \multicolumn{1}{l}{\textbf{Line F1}} \\ \midrule
Base Strong                    & 76.62                                       & 76.79                                    & 76.68                                & 76.62                                         & 76.79                                      & 76.68                                  & 45.53                                       & 37.83                                    & 39.31                                \\
Base Weak                      & 78.65                                       & 78.82                                    & 78.7                                 & 78.65                                         & 78.82                                      & 78.7                                   & 35.92                                       & 28.49                                    & 29.97                                \\
Self-Consistency - Direct      & \cellcolor[HTML]{F4CCCC}76.25               & \cellcolor[HTML]{F4CCCC}76.25            & \cellcolor[HTML]{F4CCCC}76.25        & \cellcolor[HTML]{F4CCCC}76.25                 & \cellcolor[HTML]{F4CCCC}76.25              & \cellcolor[HTML]{F4CCCC}76.25          & \cellcolor[HTML]{F4CCCC}35.44               & \cellcolor[HTML]{F4CCCC}28.03            & \cellcolor[HTML]{F4CCCC}29.36        \\
\bottomrule
\end{tabular}%
}
\caption{Localization performance for O3-mini as strong LM and GPT-4o-mini as weak LM. Cells highlighted in red depict lower score than the base weak LM setting.}
\label{tab:localization_o3-mini_gpt-4o-mini}
\end{table*}
\begin{table*}[!h]
\centering
\resizebox{\textwidth}{!}{%
\begin{tabular}{@{}lrrrrrrrrr@{}}
\toprule
\textbf{Method}                & \multicolumn{1}{l}{\textbf{File Precision}} & \multicolumn{1}{l}{\textbf{File Recall}} & \multicolumn{1}{l}{\textbf{File F1}} & \multicolumn{1}{l}{\textbf{Module Precision}} & \multicolumn{1}{l}{\textbf{Module Recall}} & \multicolumn{1}{l}{\textbf{Module F1}} & \multicolumn{1}{l}{\textbf{Line Precision}} & \multicolumn{1}{l}{\textbf{Line Recall}} & \multicolumn{1}{l}{\textbf{Line F1}} \\ \midrule
Base Strong                    & 76.62                                       & 76.79                                    & 76.68                                & 76.62                                         & 76.79                                      & 76.68                                  & 45.53                                       & 37.83                                    & 39.31                                \\
Base Weak                      & 76.11                                       & 76.11                                    & 76.11                                & 76.11                                         & 76.11                                      & 76.11                                  & 24.56                                       & 19.66                                    & 19.95                                \\
Self-Consistency - Direct      & \cellcolor[HTML]{F4CCCC}71.48               & \cellcolor[HTML]{F4CCCC}71.48            & \cellcolor[HTML]{F4CCCC}71.48        & \cellcolor[HTML]{F4CCCC}71.48                 & \cellcolor[HTML]{F4CCCC}71.48              & \cellcolor[HTML]{F4CCCC}71.48          & \cellcolor[HTML]{F4CCCC}23.65               & \cellcolor[HTML]{F4CCCC}17.86            & \cellcolor[HTML]{F4CCCC}18.77        \\
\bottomrule
\end{tabular}%
}
\caption{Localization performance for O3-mini as strong LM and Qwen2.5-Coder-7B as weak LM. Cells highlighted in red depict lower score than the base weak LM setting.}
\label{tab:localization_o3-mini_qwen7b}
\end{table*}
\begin{table*}[!h]
\centering
\resizebox{\textwidth}{!}{%
\begin{tabular}{@{}lrrrrrrrrr@{}}
\toprule
\textbf{Method}                & \multicolumn{1}{l}{\textbf{File Precision}} & \multicolumn{1}{l}{\textbf{File Recall}} & \multicolumn{1}{l}{\textbf{File F1}} & \multicolumn{1}{l}{\textbf{Module Precision}} & \multicolumn{1}{l}{\textbf{Module Recall}} & \multicolumn{1}{l}{\textbf{Module F1}} & \multicolumn{1}{l}{\textbf{Line Precision}} & \multicolumn{1}{l}{\textbf{Line Recall}} & \multicolumn{1}{l}{\textbf{Line F1}} \\ \midrule
Base Strong                    & 76.62                                       & 76.79                                    & 76.68                                & 76.62                                         & 76.79                                      & 76.68                                  & 45.53                                       & 37.83                                    & 39.31                                \\
Base Weak                      & 74.67                                       & 75.8                                     & 75.03                                & 74.67                                         & 75.8                                       & 75.03                                  & 37.41                                       & 30.85                                    & 32.01                                \\
Self-Consistency - Direct      & 76.35                                       & 76.35                                    & 76.35                                & 76.35                                         & 76.35                                      & 76.35                                  & 38.41                                       & \cellcolor[HTML]{F4CCCC}29.98            & \cellcolor[HTML]{F4CCCC}32           \\
\bottomrule
\end{tabular}%
}
\caption{Localization performance for O3-mini as strong LM and Qwen2.5-Coder-14B as weak LM. Cells highlighted in red depict lower score than the base weak LM setting.}
\label{tab:localization_o3-mini_qwen14b}
\end{table*}
\begin{table*}[!h]
\centering
\resizebox{\textwidth}{!}{%
\begin{tabular}{@{}lrrrrrrrrr@{}}
\toprule
\textbf{Method}                & \multicolumn{1}{l}{\textbf{File Precision}} & \multicolumn{1}{l}{\textbf{File Recall}} & \multicolumn{1}{l}{\textbf{File F1}} & \multicolumn{1}{l}{\textbf{Module Precision}} & \multicolumn{1}{l}{\textbf{Module Recall}} & \multicolumn{1}{l}{\textbf{Module F1}} & \multicolumn{1}{l}{\textbf{Line Precision}} & \multicolumn{1}{l}{\textbf{Line Recall}} & \multicolumn{1}{l}{\textbf{Line F1}} \\ \midrule
Base Strong                    & 76.62                                       & 76.79                                    & 76.68                                & 76.62                                         & 76.79                                      & 76.68                                  & 45.53                                       & 37.83                                    & 39.31                                \\
Base Weak                      & 76.9                                        & 78.28                                    & 77.36                                & 76.9                                          & 78.28                                      & 77.36                                  & 41.65                                       & 34.76                                    & 35.46                                \\
Self-Consistency - Direct      & 78.63                                       & 79.73                                    & 78.96                                & 78.63                                         & 79.73                                      & 78.96                                  & 42.58                                       & \cellcolor[HTML]{F4CCCC}34.75            & 35.76                                \\
\bottomrule
\end{tabular}%
}
\caption{Localization performance for O3-mini as strong LM and Qwen2.5-Coder-32B as weak LM. Cells highlighted in red depict lower score than the base weak LM setting.}
\label{tab:localization_o3-mini_qwen32b}
\end{table*}
\begin{table*}[!h]
\centering
\resizebox{\textwidth}{!}{%
\begin{tabular}{@{}lrrrrrrrrr@{}}
\toprule
\textbf{Method}                & \multicolumn{1}{l}{\textbf{File Precision}} & \multicolumn{1}{l}{\textbf{File Recall}} & \multicolumn{1}{l}{\textbf{File F1}} & \multicolumn{1}{l}{\textbf{Module Precision}} & \multicolumn{1}{l}{\textbf{Module Recall}} & \multicolumn{1}{l}{\textbf{Module F1}} & \multicolumn{1}{l}{\textbf{Line Precision}} & \multicolumn{1}{l}{\textbf{Line Recall}} & \multicolumn{1}{l}{\textbf{Line F1}} \\ \midrule
Base Strong                    & 78.65                                       & 78.82                                    & 78.7                                 & 78.65                                         & 78.82                                      & 78.7                                   & 35.92                                       & 28.49                                    & 29.97                                \\
Base Weak                      & 76.11                                       & 76.11                                    & 76.11                                & 76.11                                         & 76.11                                      & 76.11                                  & 24.56                                       & 19.66                                    & 19.95                                \\
Self-Consistency - Direct      & \cellcolor[HTML]{F4CCCC}71.48               & \cellcolor[HTML]{F4CCCC}71.48            & \cellcolor[HTML]{F4CCCC}71.48        & \cellcolor[HTML]{F4CCCC}71.48                 & \cellcolor[HTML]{F4CCCC}71.48              & \cellcolor[HTML]{F4CCCC}71.48          & \cellcolor[HTML]{F4CCCC}23.65               & \cellcolor[HTML]{F4CCCC}17.86            & \cellcolor[HTML]{F4CCCC}18.77        \\
\bottomrule
\end{tabular}%
}
\caption{Localization performance for GPT-4o-mini as strong LM and Qwen2.5-Coder-7B as weak LM. Cells highlighted in red depict lower score than the base weak LM setting.}
\label{tab:localization_gpt-4o-mini_qwen7b}
\end{table*}

\end{document}